\documentclass[letterpaper]{article} 
\usepackage{aaai2026}  
\usepackage{times}  
\usepackage{helvet}  
\usepackage{courier}  
\usepackage[hyphens]{url}  
\usepackage{graphicx} 
\def\UrlFont{\rm}  
\usepackage{natbib}  
\usepackage{caption} 
\usepackage{algorithm}
\usepackage{algorithmic}
\usepackage{amsmath}
\usepackage{booktabs}
\usepackage{xcolor}
\usepackage{caption}
\usepackage{enumitem}
\usepackage{array}
\usepackage[capitalise,noabbrev]{cleveref}
\usepackage{multibib}
\newcites{appendix}{appendix}

\usepackage{newfloat}
\usepackage{listings}
\DeclareCaptionStyle{ruled}{labelfont=normalfont,labelsep=colon,strut=off} 
\floatstyle{ruled}
\newfloat{listing}{tb}{lst}{}
\floatname{listing}{Listing}
\title{MULTIBENCH++: A Unified and Comprehensive Multimodal Fusion Benchmarking Across Specialized Domains}

\author {
    Leyan Xue\textsuperscript{\rm 1},
    Changqing Zhang\textsuperscript{\rm 1}\thanks{Corresponding author.}, 
    Kecheng Xue\textsuperscript{\rm 2},
    Xiaohong Liu\textsuperscript{\rm 3},
    Guangyu Wang\textsuperscript{\rm 2},
    Zongbo Han\textsuperscript{\rm 2}\footnotemark[1] 
    \\
}
\affiliations {
    \textsuperscript{\rm 1}College of Intelligence and Computing, Tianjin University, Tianjin, China\\
    \textsuperscript{\rm 2}State Key Laboratory of Networking and Switching Technology, Beijing University of Posts and Telecommunications, Beijing, China\\
    \textsuperscript{\rm 3}Institute of Medical Artificial Intelligence, South China Hospital, Medical School, Shenzhen University, Guangdong, China
    \\
}

\usepackage{bibentry}
\usepackage{cuted}
\usepackage{placeins}
\usepackage[toc,page]{appendix}
\usepackage{makecell} 
\usepackage{multirow}
\begin{document}

\maketitle


\begin{abstract}
\textcolor{black}{Although multimodal fusion has made significant progress, its advancement is severely hindered by the lack of adequate evaluation benchmarks. Current fusion methods are typically evaluated on a small selection of public datasets, a limited scope that inadequately represents the complexity and diversity of real-world scenarios, potentially leading to biased evaluations.
This issue presents a twofold challenge. On one hand, models may overfit to the biases of specific datasets, hindering their generalization to broader practical applications. On the other hand, the absence of a unified evaluation standard makes fair and objective comparisons between different fusion methods difficult. Consequently, a truly universal and high-performance fusion model has yet to emerge.
To address these challenges, we have developed a large-scale, domain-adaptive benchmark for multimodal evaluation. This benchmark integrates over 30 datasets, encompassing 15 modalities and 20 predictive tasks across key application domains. To complement this, we have also developed an open-source, unified, and automated evaluation pipeline that includes standardized implementations of state-of-the-art models and diverse fusion paradigms.
Leveraging this platform, we have conducted large-scale experiments, successfully establishing new performance baselines across multiple tasks. This work provides the academic community with a crucial platform for rigorous and reproducible assessment of multimodal models, aiming to propel the field of multimodal artificial intelligence to new heights.} 
\end{abstract}


\begin{links}
    \link{Code}{https://github.com/ravexly/MultiBenchplus}
    \link{Project Page}{https://ravexly.github.io/MultiBenchplus}
\end{links}

\section{Introduction}

\begin{table*}[h!]
\centering
\caption{MULTIBENCH++ offers a unified benchmark suite of 37 multimodal datasets spanning a wide spectrum of research fields, data scales, input modalities( $a$: audio, $c$: clinical/tabular, ${d}$: depth/DSM, $e$: events/spiking, ${f}$: 2D fundus, ${g}$: GIS, ${h}$: HSI, ${i}$: image, $k$: time-series, ${L}$: LiDAR, ${m}$: multispectral, ${M}$: metadata, ${o}$: multi-omics, ${O}$: 3D OCT, ${s}$: SAR, ${t}$: text, ${v}$: video, with omics sub-typed as ${o_{\!1}}$: mRNA, ${o_{\!2}}$: miRNA, ${o_{\!3}}$: DNA), and downstream tasks.}
\resizebox{\textwidth}{!}{%
\label{tab:my_styled_table}
\begin{tabular}{lllll}
\toprule
\textbf{Domain} & \textbf{Dataset} & \textbf{Modalities} & \textbf{\# Samples} & \textbf{Prediction Task} \\
\midrule
\multirow{6}{*}{\shortstack[l]{Remote Sensing}} & Houston2013 & $\{h, L\}$ & 14,999 & Land Cover Classification \\
& Houston2018 & $\{h, L\}$ & 2,018,910 & Land Cover Classification \\
& MUUFL Gulfport & $\{h, L\}$ & 53,687 & Land Cover Classification \\
& Trento & $\{h, L\}$ & 30,214 & Land Cover Classification \\
& Berlin & $\{h, s\}$ & 464,671 & Land Cover Classification \\
& MDAS (Augsburg) & $\{ h, s\}$ & 78,294 & Land Cover Classification \\
& ForestNet & $\{c, i , m\}$ & 2,757 & Forest Type Mapping \\
\hline
\multirow{8}{*}{\shortstack[l]{Medical AI}} & TCGA-BRCA & $\{o_1,o_2,o_3\}$ & 875 & Survival Prediction, Subtype Classification \\
& ROSMAP & $\{o_1,o_2,o_3\}$ & 351 & Disease Progression Prediction \\
& SIIM-ISIC & $\{i, M\}$ & 33,126 & Malignant Tumor Classification \\
& Derm7pt & $\{i, M\}$ & 1,011 & Lesion Diagnosis Prediction \\
& GAMMA & $\{f, O\}$ & 100 & Glaucoma Grading \\
& MIMIC-III & $\{c,t\}$ & 36,212 & Mortality Prediction \\
& MIMIC-CXR & $\{c,t\}$ & 372,147 & Mortality Prediction, Multilabel Classification\\
& eICU & $\{c,t\}$ & 7,637 & Mortality Prediction \\
& TCGA & $\{o_1,o_2,o_3\}$ & 306 & Subtype classification, Tumor Malignancy Grading \\
\hline
\multirow{12}{*}{\shortstack[l]{Affective Computing\\\&Social Media\\Understanding}} & MELD & $\{a, t\}$ & 13,708 & Emotion/Sentiment Recognition \\
& IEMOCAP & $\{a, v, t\}$ & 7,433 & Emotion/Sentiment Recognition \\

& MAMI & $\{i, t\}$ & 11,000 & Misogyny Content Detection \\
& Memotion & $\{i, t\}$ & 6,831 & Offensive Content Detection \\
& MUTE & $\{i, t\}$ & 4,156 & Hate Speech Detection \\
& MultiOFF & $\{i, t\}$ & 743 & Offensive Content Detection \\
& MET-Meme(C) & $\{i, t\}$ & 2,299 & Metaphor/Emotion/Intent Recognition \\
& MET-Meme(E) & $\{i, t\}$ & 1,053 & Metaphor/Emotion/Intent Recognition \\
& CH-SIMS & $\{a, v, t\}$ & 2,281 & Sentiment Analysis \\
& CH-SIMS v2.0 & $\{a, v, t\}$ & 4,403 & Sentiment Analysis \\
& Twitter2015 & $\{i, t\}$ & 5,338 & Multimodal Named Entity Recognition \\
& Twitter1517 & $\{i, t\}$ & 4,672 & Multimodal Named Entity Recognition \\
\hline
\multirow{9}{*}{\shortstack[l]{Others}} & MIRFLICKR & $\{i, t\}$ & 20,015 & Image Retrieval \\
& CUB Image-Caption & $\{i, t\}$ & 117,880 & Fine-grained Classification \\
& SUN-RGBD & $\{i, d\}$ & 9,504 & Scene Understanding, Object Detection \\
& NYUDv2 & $\{i, d\}$ & 1,863 & Scene Understanding, Object Detection \\
& UPMC-Food101 & $\{i, t\}$ & 90,686 & Food Recognition \\
& MVSA-Single & $\{i, t\}$ & 2,592 & Sentiment Analysis\\
& MNIST-SVHN & $\{i, i\}$ & 660,680 & Digit Recognition \\
& N-MNIST+N-TIDIGITS & $\{e, i\}$ & 4,050 & Digit Recognition \\
& E-MNIST+EEG & $\{i, k\}$ & 702 & Digit Recognition \\
\bottomrule
\end{tabular}
}
\end{table*}

\textcolor{black}{Multimodal data, such as text, images, and sensor signals, is driving the next generation of artificial intelligence. Through a technique known as Multimodal Fusion, AI systems can integrate and understand information from these diverse sources, achieving a more comprehensive, accurate, and robust understanding than is possible with any single source \citep{baltruvsaitis2018multimodal,xu2023multimodal}. This capability is a key driver for advancing AI to higher levels of intelligence and shows immense potential in fields like autonomous driving and medical diagnostics \citep{caesar2020nuscenes,azam2022review}.}

However, a significant divergence exists between multimodal research and other domains. While fields such as natural language processing and graph learning have successfully converged on dominant architectural paradigms, specifically the Transformer and GNNs, the multimodal domain conspicuously lacks an equivalent unified, foundational framework. Progress remains highly fragmented; researchers typically validate new methods on a small, bespoke selection of classic datasets. This reliance on siloed benchmarks is a key bottleneck. It not only leads to models overfitting to specific data biases and prevents fair objective comparisons, but more importantly, it has hindered the systematic search for a truly general-purpose fusion architecture. Four years ago, the groundbreaking MULTIBENCH framework partially addressed this by providing a unified evaluation platform. Yet, with the field's rapid evolution, its limitations are now apparent, and it is no longer sufficient to meet today's challenges.

The urgent need for a new foundational platform stems from two primary trends. The first is the explosive growth in data combinatorial complexity. Unlike unimodal tasks, real-world applications in medical imaging, IoT, and autonomous driving \citep{hu2023mdas, kong2011integrative} span a vast, heterogeneous spectrum. This combinatorial effect, where different data combinations can yield entirely different analytical conclusions, means older, simpler benchmarks can no longer demonstrate a model's robustness or adaptability to real-world complexity. The second trend is the rapid evolution of fusion models, especially Transformer-based methods \citep{wei2020multi, wang2022multimodal, xu2023multimodal}. Without a standardized, complex testbed, it is impossible to determine if these new models are truly general-purpose or simply adept at specific data combinations. Therefore, a platform that forces models to confront this combinatorial complexity has become essential to guide the search for a foundational architecture.

\textcolor{black}{To address key challenges in evaluating next-generation multimodal fusion models, we introduce MULTIBENCH++, a new, large-scale benchmark. Instead of being a simple incremental update, MULTIBENCH++ represents a significant leap forward in terms of scale, domain diversity, and suitability for modern architectures. Its core contributions are threefold:
\begin{itemize}
    \item Expanded scale and domain coverage. MULTIBENCH++ brings together over 30 datasets, more than doubling the size of its predecessor. More importantly, it extends into highly complex and specialized domains, including Remote Sensing, Healthcare, Affective Computing, and Social Media Analysis. These domains present unique data fusion challenges.
    \item Designed for rigorous testing of advanced architectures. Its datasets are carefully chosen for their high complexity, rich interplay between modalities, and naturally occurring missing data, creating a challenging test environment. This design allows for rigorous testing of advanced, Transformer-based architectures and novel fusion techniques.
    \item An open-source framework for robust and reproducible evaluation. To ensure fair and rigorous scientific comparisons, MULTIBENCH++ includes a standardized, open-source evaluation framework. This framework provides standardized data splits, Robustness Probes, and a set of strong baseline models that have been carefully tuned using Automated Hyperparameter Optimization. This infrastructure is designed to lower the barrier to entry for researchers and ensure that future innovations can be reliably evaluated on a fair and consistent foundation.
\end{itemize}}

\begin{figure*}[htbp]
\centering
\includegraphics[width=1\textwidth]{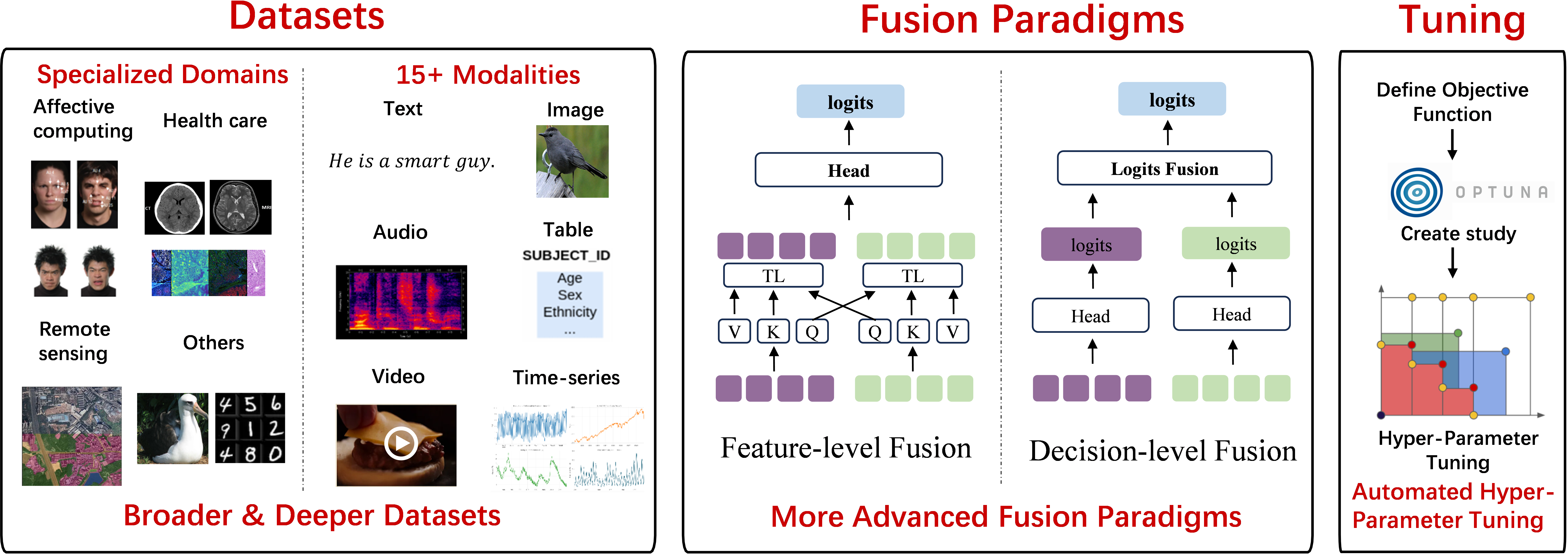} 
\caption{ An overview of the MULTIBENCH++ framework, highlighting our core contributions. (Left) We introduce a broader and deeper collection of datasets, significantly expanding into more specialized domains and data modalities. (Center) We integrate more advanced fusion paradigms, including feature-level transformer-based fusion and decision-level fusion. (Right) We provide an automated hyper-parameter tuning platform, powered by Optuna, to ensure robust and reproducible evaluation.}
\label{fig:moti}
\end{figure*}
\section{Related Works}
\paragraph{Comparisons with Related Benchmarks.}
Multimodal research has been driven by a series of influential benchmarks. 
Foundational datasets for visual question answering, such as VQA-v2 \citep{goyal2017making}, established large-scale, open-ended visual reasoning as a core challenge. 
In parallel, benchmarks for multimodal sentiment analysis and emotion recognition, like CMU-MOSI \citep{zadeh2016mosi} and the larger CMU-MOSEI \citep{zadeh2018multimodal}, provided key testbeds for integrating language, visual, and acoustic signals.
To standardize evaluation across this growing landscape, frameworks like MULTIBENCH \citep{liang2021multibench} were introduced to offer a unified, reproducible testbed for assessing model robustness across a diverse set of tasks. 
Building on this principle, our work expands this suite with 20 additional datasets, contributing to a broader trend of comprehensive evaluation that also includes integrating more competitive methods and developing automated tuning platforms. \\
Other works introduce new domains, such as MM-GRAPH \citep{zhu2025mosaic}, which integrates visual features into graph-based tasks, and Dyn-VQA \citep{libenchmarking}, which tests dynamic question answering requiring multi-hop retrieval.

\paragraph{Multimodal Fusion.}
The core challenge in multimodal learning is fusion: the effective combination of information from different modalities. 
Classical approaches are categorized by the architectural stage at which fusion occurs: early (feature-level) fusion concatenates raw or low-level features \citep{ramachandram2017deep,atrey2010multimodal}, while late (decision-level) fusion combines outputs from modality-specific models \citep{soleymani2017survey,han2022trusted,zhang2023provable}.
The advent of the Transformer has made attention-based fusion the dominant paradigm \citep{xu2023multimodal}. 
These methods can be broadly classified as single-stream, where multimodal inputs are concatenated and processed by a unified encoder, or multi-stream, which uses separate encoders for each modality followed by cross-attention mechanisms to integrate information \citep{nagrani2021attention}.
Such architectures allow for nuanced, dynamically-weighted integration of modal features. 
More recent research also focuses on developing fusion techniques that are robust to real-world challenges like noisy, incomplete, or imbalanced data \citep{zhang2024multimodal}.

\paragraph{Analysis of Multimodal Representations.}
Beyond predictive performance, understanding the quality of learned multimodal representations is a critical area of research \citep{frome2013devise,kiros2014unifying}.
A primary technique is the use of probing tasks, where simple classifiers are trained on a model's internal embeddings to test for specific encoded properties \citep{alain2016understanding}. 
For example, studies have shown that global semantics are often encoded in the intermediate layers of MLLMs, while local, fine-grained details are captured in the final layers \citep{tao2024probing}.
Other works focus on interpretability and explainability, using techniques like visualizing attention maps or model dissection tools to attribute a model's decision to specific unimodal features \citep{liang2024foundations}.
The overarching goal of these analysis is to measure fundamental properties like modality alignment, complementarity, and redundancy, which is crucial for building more robust and efficient models .
\section{MULTIBENCH++ :  A Broader \& Deeper Multimodal Benchmark}

\begin{algorithm*}
    \caption{End-to-End Workflow with Optuna.}
    \label{algo:pipe}
    \begin{lstlisting}[language=Python]
# 1. Load data and define models
train_loader, val_loader, test_loader, n_classes = get_loader()
encoders = [Modality1Encoder(args), Modality2Encoder(args)]
fusion = TMC(n_classes)
head = get_head(n_classes, decision=True)
# 2. Define a minimal objective function
def objective(trial):
    # Suggest hyperparameters
    params = {'lr': trial.suggest_loguniform('lr', 1e-5, 1e-3), ...}
    # Train a model and return its validation accuracy
    val_accuracy = train(encoders, fusion, head, params, ...)
    return val_accuracy
# 3. Run the hyperparameter search
study = optuna.create_study(direction='maximize')
study.optimize(objective, n_trials=10)
# 4. Get the best parameters and build the final model
final_model = build_model(study.best_trial.number)
# 5. Evaluate the final, optimized model
test_accuracy = test(final_model, test_loader)
    \end{lstlisting}
\end{algorithm*}
\subsection{Background}
Multimodal datasets differ in both the types of information they provide and the ways that information is encoded. Remote sensing archives couple spectral bands with LiDAR point clouds, while electronic health records weave free-text notes together with structured lab values and pathology slides. Affective repositories, in turn, align video frames to audio streams and wearable signals. These collections are rarely designed for joint use, so their formats, resolutions, and noise levels diverge. A benchmark must therefore expose models to this heterogeneity. As shown in Fig. \ref{fig:moti}, MULTIBENCH++ addresses the current evaluation gap by collecting over thirty datasets drawn from highly specialized domains including remote sensing, healthcare, affective computing and social media. \textcolor{black}{
These specialized domains are not arbitrary. they represent critical frontiers where multimodal integration is pivotal for scientific and societal progress. The inclusion of such a wide array of data sources ensures that the benchmark rigorously tests a model's ability to generalize across fundamentally different data structures and noise profiles, moving beyond single-domain evaluations.
}
\subsection{Datasets}

\subsubsection{Remote Sensing for Environmental Intelligence}
The remote-sensing domain for environmental intelligence tackles land-cover classification, target detection, spectral unmixing, and related tasks by fusing data whose physical origins differ fundamentally. Optical and hyperspectral systems capture surface chemistry yet remain weather-dependent; SAR measures microwave backscatter day-and-night; LiDAR delivers centimetre-level topography and canopy structure. Integrating these streams requires reconciling disparate spatial resolutions, geometries, and noise statistics.
Foundational datasets establish this task-oriented landscape. Houston2013 and Houston2018 \citep{debes2014hyperspectral,xu2018hyperspectral} combine hyperspectral imagery with LiDAR for urban land-cover classification. MUUFL Gulfport \citep{gader2013muufl} adds co-registered hyperspectral and LiDAR data over a university campus for classification and rare-target detection. Trento provides a rural counterpart with hyperspectral and LiDAR. Berlin \citep{okujeni2016berlin} fuses PolSAR and hyperspectral data, while ForestNet \citep{irvin2020forestnet} couples satellite imagery and airborne LiDAR for forest-type and biomass mapping. The recent MDAS dataset \citep{hu2023mdas} enriches the benchmark suite with simultaneous SAR, multispectral, hyperspectral, DSM, and GIS layers, supporting resolution enhancement, unmixing, and classification. Together, these datasets constitute a systematic test-bed for advancing theoretically grounded and practically robust environmental-intelligence algorithms.

\subsubsection{Medical AI for Diagnostics and Prognosis}
The medical-intelligence domain addresses survival prediction, malignancy classification, disease-progression modelling and related tasks by fusing exceptionally heterogeneous data streams. Gigapixel whole-slide images quantify tissue morphology; high-dimensional omics profiles capture molecular aberrations; dense time-series vital signs and concise clinical narratives encode patient trajectories. Integrating these modalities demands reconciling extreme differences in resolution, scale and noise, while preserving clinical interpretability.
Benchmark datasets anchor this landscape. TCGA-BRCA \citep{weinstein2013cancer} couples WSIs with multi-omics for breast-cancer survival and subtype analysis. ROSMAP \citep{bennett2018religious} supplies longitudinal multi-omics and pathology to chart Alzheimer progression. In dermatology, SIIM-ISIC \citep{rotemberg2021patient} and Derm7pt \citep{kawahara2018seven} pair dermoscopic images with patient metadata for melanoma detection. GAMMA \citep{wu2023gamma} fuses 2D fundus photographs with 3D OCT volumes for glaucoma grading and optic-disc/cup segmentation. MIMIC-III \citep{johnson2016mimic}, MIMIC-CXR \cite{johnson2019mimic} and eICU \cite{pollard2018eicu} deliver large-scale ICU time series merged with static clinical records to support mortality prediction and disease-code classification.

\subsubsection{Affective Computing and Social Media Understanding}
The Affective Computing domain addresses emotion recognition, sarcasm detection,  sentiment analysis and related high-level tasks by fusing text, acoustic, visual and cultural cues that are frequently incongruent or metaphorical. Dyadic and multi-party conversations introduce temporal alignment challenges, while internet memes overlay visual symbols with rapidly shifting socio-cultural contexts. Effective integration demands models that can resolve cross-modal sarcasm, capture long-range conversational flow and remain sensitive to cultural nuance.
Benchmark datasets collectively span these phenomena. MELD \citep{poria2019meld} and IEMOCAP \citep{busso2008iemocap} provide temporally aligned audio, video and transcriptions for multi-party and dyadic emotion recognition. MUTE \citep{hossain2022mute} extends the conversational setting to multilingual scenarios. MAMI \citep{fersini2022semeval}, MultiOFF \citep{suryawanshi2020multimodal}, Memotion \citep{sharma2020semeval} and MET-Meme \cite{xu2022met} (Chinese \& English versions) jointly encode images and text for the detection of misogynistic, offensive and metaphorical content in memes. 
CH-SIMS \citep{yu2020ch} and its successor CH-SIMS v2.0 \citep{liu2022make} deliver fine-grained Chinese multimodal sentiment annotations with explicit modality-importance scores. Twitter2015 and Twitter1517 \citep{zhang2018adaptive,lu2018visual,chen2023HFIR} close the loop with classic social-media tasks, linking text and images for named-entity recognition and sentiment polarity prediction.
\subsubsection{Others}
This domain addresses image–text retrieval, fine-grained classification, digit recognition, scene understanding and related tasks by fusing heterogeneous yet tightly aligned modalities. Integrating vision with language, depth, audio or event streams demands reconciling distinct resolutions, sampling rates and noise distributions while preserving interpretability.
Benchmark datasets anchor this landscape. MIRFLICKR-25K \cite{huiskes2008mir} couples images with user tags for large-scale retrieval. MVSA-Single \citep{niu2016sentiment} supplies tweet images and text for visual–sentiment classification. CUB Image-Caption \citep{shi2019variational} pairs bird photographs with textual descriptions for fine-grained classification, while MNIST-SVHN \citep{shi2019variational} aligns handwritten and street-view digits across domains. SUN-RGBD \citep{song2015sun} and NYUDv2 \citep{silberman2012indoor} provide RGB–depth pairs for scene understanding and object detection, and UPMC-Food101 \citep{wang2015recipe} fuses food images with recipe text for cross-modal recognition. Following \citet{lin2025resistive}, we further combine N-MNIST+N-TIDIGITS \citep{orchard2015converting,anumula2018feature} to synchronise frame-based and event-based vision with spoken digits, and E-MNIST+EEG \citep{cohen2017emnist,willett2021high} to link character images with electroencephalography signals for cognitive-state-aware digit recognition.
\subsection{Evaluation Protocol}
We follow MULTIBENCH’s holistic evaluation with only minor adjustments.
For every dataset and method, we report performance on the test fold using task-specific metrics (e.g., accuracy, macro-F1, AUPRC, or MSE).
%
\textcolor{black}{Each run is performed 3 times using different random seeds, all under the same hardware configuration.}

\begin{table*}[htbp]
  \centering
  \caption{Performance on Remote Sensing Datasets.}
  \label{tab:rs}
  \resizebox{\textwidth}{!}{%
  \begin{tabular}{@{}l *{11}{>{\centering\arraybackslash}p{1.2cm}} @{}}
    \toprule
    \textbf{Dataset} & 
    \textbf{Concat} & 
    \textbf{TF} & 
    \textbf{Concat Early} & 
    \textbf{LFT} & 
    \textbf{EFT} & 
    \textbf{Multi-to-One} & 
    \textbf{One-to-Multi} & 
    \textbf{CAF} & 
    \textbf{CACF} & 
    \textbf{LS} & 
    \textbf{TMC} \\
    \midrule
    Houston2013 & 76.36 & 79.72 & 79.59 & 69.27 & 21.60 & 78.56 & 79.97 & 74.12 & 83.66 & 80.55 & 81.52 \\
    Houston2018 & 79.30 & 79.24 & 78.52 & 66.30 & 48.58 & 71.34 & 74.52 & 75.99 & 82.20 & 78.09 & 78.52 \\
    Berlin & 79.03 & 74.45 & 79.47 & 64.23 & 62.63 & 69.95 & 68.75 & 75.01 & 78.26 & 78.91 & 77.93 \\
    MUUFL Gulfport & 84.21 & 83.90 & 86.26 & 71.77 & 46.68 & 80.95 & 81.23 & 83.99 & 86.42 & 86.64 & 83.20 \\
    Trento & 98.43 & 96.95 & 98.14 & 95.68 & 71.64 & 97.71 & 96.08 & 97.74 & 98.68 & 98.53 & 97.67 \\
    Augsburg & 89.24 & 85.86 & 89.50 & 82.88 & 57.29 & 85.86 & 86.80 & 87.92 & 89.05 & 89.49 & 87.21 \\
    ForestNet & 45.18 & 45.63 & 45.68 & 47.19 & 44.33 & 45.78 & 45.58 & 45.03 & 45.93 & 46.08 & 45.68 \\
    \bottomrule
  \end{tabular}
  }
\end{table*}
\begin{table*}[htbp]
  \centering
  \caption{Performance on Medical AI Datasets. The dash ``-" indicates the method is not applicable to this dataset.}
  \label{tab:med}
  \resizebox{1\textwidth}{!}{
  \begin{tabular}{@{}l *{11}{>{\centering\arraybackslash}p{1.2cm}} @{}}
    \toprule
    \textbf{Dataset} & 
    \textbf{Concat} & 
    \textbf{TF} & 
    \textbf{Concat Early} & 
    \textbf{LFT} & 
    \textbf{EFT} & 
    \textbf{Multi-to-One} & 
    \textbf{One-to-Multi} & 
    \textbf{CAF} & 
    \textbf{CACF} & 
    \textbf{LS} & 
    \textbf{TMC} \\
    \midrule
    TCGA-BRCA & 78.17  & 77.18  & 77.18  & 69.44  & 65.67  & 76.79  & 76.19  & 78.77  & 78.37  & 77.18  & 75.40  \\
    ROSMAP & 70.95  & 75.71  & 70.00  & 42.86  & 69.52  & 69.52  & 68.10  & 66.67  & 71.90  & 71.43  & 66.67  \\
    SIM-ISIC & 97.86  & 97.77  & 97.86  & 97.83  & 97.85  & 97.85  & 97.83  & 97.86  & 97.83  & 97.83  & 97.85  \\
    Derm7pt & 45.49  & 52.72  & 45.41  & 43.96  & 38.86  & 45.92  & 46.77  & 48.13  & 52.72  & 46.34  & 52.72  \\
    GAMMA &  61.43  & 62.86  & 63.81  & 62.38  & 59.05  & 57.62  & 65.71  & 63.33  & 63.33  & 62.38  & 61.43  \\
    MIMIC-III & 68.09  & 68.81  & 68.48  & 68.42  & 68.59  & 69.01  & 68.76  & 68.51  & 68.88  & 68.47  & 68.87  \\
    eICU & 90.05  & 90.03  & 90.05  & 90.05  & 90.05  & 90.05  & 90.05  & 90.07  & 90.05  & 90.05  & 90.03  \\
    TCGA & 51.91  & - & 53.55  & 60.66  & 51.37  & 60.11  & 53.01  & 56.28  & 61.75  & 54.10  & 62.84  \\
    MIMIC-CXR (macro-F1) & 0.6861  & 0.8458  & 0.6788  & 0.1551  & 0.1829  & 0.5229  & 0.4031  & 0.6136  & 0.7523  & 0.7644 & - \\
    \bottomrule
  \end{tabular}
  }
\end{table*}
\section{MULTIBENCH++ Algorithms: More Advanced Fusion  Paradigms}
\label{algorithms}

We introduce a complete, end-to-end framework for systematic multimodal evaluation. The framework is built around two classes of fusion methodologies: four Transformer-centric paradigms to model complex cross-modal interactions, and two modules for efficient decision-level logits fusion. To enable robust and reproducible experimentation, we further develop an automated hyperparameter optimization engine based on Optuna \citep{akiba2019optuna}. This engine facilitates a systematic exploration and optimization, allowing for an efficient identification of optimal configurations.
\subsection{Transformer-Based Feature Fusion Architectures}
We evaluate several Transformer-based architectures for multimodal feature fusion. Each model is designed to accept a set of modality-specific input tensors $\{x_i\}_{i=1}^k$ and produce a unified representation vector $g$.
\subsubsection*{Hierarchical Attention (Multi-to-One)}
This model first encodes each modality independently using a shallow, modality-specific Transformer encoder $\mathcal{E}_i$. The resulting classification tokens are then concatenated and processed by a deeper, shared fusion encoder $\mathcal{E}_{\text{fuse}}$ to model high-level interactions \cite{li2021ai}.
\[
z_i = \mathcal{T}(\mathcal{E}_i(\Phi_i(x_i)))
\]
\[
g = \mathcal{T}(\mathcal{E}_{\text{fuse}}([z_1; z_2; \dots; z_k]))
\]
where $\Phi_i$ is a 1D convolution and $\mathcal{T}(\cdot)$ is an operator that selects the classification token's embedding.

\subsubsection*{Hierarchical Attention (One-to-Multi)}
Conversely, this architecture first models cross-modal interactions before refining modality-specific features. All inputs are projected by a linear layer $\Psi_i$ and concatenated into a single sequence. This joint sequence is processed by a shared encoder $\mathcal{E}_{\text{shared}}$. The output sequence is then split into its original modality-specific segments, each of which is passed through a final dedicated encoder $\mathcal{E}_i$ \cite{lin2020interbert}.
\[
h = \mathcal{E}_{\text{shared}}([\Psi_1(x_1); \dots; \Psi_k(x_k)])
\]
\[
[h_1, \dots, h_k] = \text{Split}(h)
\]
\[
g = [\mathcal{T}(\mathcal{E}_1(h_1)); \dots; \mathcal{T}(\mathcal{E}_k(h_k))]
\]

\subsubsection*{Cross-Attention Fusion (CAF)}
CAF facilitates direct, dense interaction between pairs of modalities. For a bimodal case ($x_1, x_2$), each modality's sequence is used to generate queries that attend to the keys and values of the other modality \cite{lu2019vilbert}.
\[
z_{1 \leftarrow 2} = \text{MultiHead}(\Psi_Q(x_1), \Psi_K(x_2), \Psi_V(x_2))
\]
\[
z_{2 \leftarrow 1} = \text{MultiHead}(\Psi_Q(x_2), \Psi_K(x_1), \Psi_V(x_1))
\]
\[
g = [\mathcal{T}(z_{1 \leftarrow 2}); \mathcal{T}(z_{2 \leftarrow 1})]
\]
where $\Psi_Q, \Psi_K, \Psi_V$ are modality-specific linear projection layers.

\subsubsection*{Cross-Attention Concatenation Fusion (CACF)}
CACF extends CAF by incorporating an additional global reasoning step \citep{zhan2021product1m,tsai2019multimodal}. The cross-attended representations ($z_{1 \leftarrow 2}, z_{2 \leftarrow 1}$) are concatenated with initial linear projections of the original inputs ($x'_i = \Psi_i(x_i)$). This combined sequence is then processed by a final global Transformer encoder $\mathcal{E}_{\text{global}}$ .
\[
f = [x'_1; z_{1 \leftarrow 2}; x'_2; z_{2 \leftarrow 1}]
\]
\[
g = \mathcal{T}(\mathcal{E}_{\text{global}}(f))
\]

\subsection*{Hybrid Logit Fusion Methods}
We also implement two representative methods that operate directly on the output logits $\{\ell_i\}_{i=1}^k$ from modality-specific classifiers. Other methods can also be easily and quickly incorporated into our proposed benchmark.

\subsubsection*{Logit Summation (LS)}
This is the most direct parameter-free method for logit fusion. It operates under the assumption that each modality contributes equally to the final prediction. The logit vectors from all modality-specific classifiers are simply summed to produce the final fused logits: \(
\ell_{\text{fused}} = \sum_{i=1}^k \ell_i
\).

\subsubsection*{Evidential Fusion (TMC)}
This method, based on evidential deep learning, transforms logits into evidence parameters $\alpha_i$ for a Dirichlet distribution \citep{han2022trusted}. This allows for explicit uncertainty quantification. The evidence from each modality is then fused using Dempster's rule of combination.
\[
\alpha_i = \text{Softplus}(\ell_i) + 1, \alpha_{\text{fused}} = \bigoplus_{i=1}^k \alpha_i
\]
Here, $\bigoplus$ denotes the Dempster-Shafer combination operator. The final class probabilities are derived from the fused evidence vector $\alpha_{\text{fused}}$.

\subsection{Automated Hyper-Parameter Tuning with Optuna}
MULTIBENCH++ radically simplifies hyper-parameter tuning by using \textbf{Optuna}, eliminating traditional, GPU-heavy grid searches.
A single \texttt{objective(trial)} callback efficiently handles the entire process, including:
\begin{itemize}
    \item Dynamic search-space definition
    \item Module re-instantiation
    \item Early-stopping pruning
    \item Best checkpoint storage
\end{itemize}
This automated approach drastically cuts tuning time and resources, significantly boosting efficiency and performance.
\paragraph{Search-Space Specification}
For every trial, Optuna independently samples
\begin{itemize}
  \item learning rate $\log\mathcal{U}(10^{-5}, 10^{-3})$,
  \item weight decay $\log\mathcal{U}(10^{-6}, 10^{-2})$,
  \item optimizer type $\in \{\text{AdamW}, \text{RMSprop}, \text{Adam}\}$.
\end{itemize}

Thus, the joint space spans three orders of magnitude in learning dynamics and two architectural regimes, while remaining compact for efficient Bayesian optimisation.
\paragraph{End-to-End Workflow}


Algorithm \ref{algo:pipe} demonstrates the end-to-end workflow.
After retrieving a dataset via the unchanged data loader, one may substitute any of the presented Transformer fusion modules or logits-level combiners; the Optuna wrapper then orchestrates the hyper-parameter search and returns a trained model, which is subsequently evaluated under the standard protocol.
\begin{table*}[htbp]
  \centering
  \caption{Performance on Affective Computing \& Social Media Understanding Datasets. The dash ``-" indicates the method is not applicable to this dataset.}
  \label{tab:ac}
  \resizebox{\textwidth}{!}{%
  \begin{tabular}{@{}l *{11}{>{\centering\arraybackslash}p{1.2cm}} @{}}
    \toprule
    \textbf{Dataset} & 
    \textbf{Concat} & 
    \textbf{TF} & 
    \textbf{Concat Early} & 
    \textbf{LFT} & 
    \textbf{EFT} & 
    \textbf{Multi-to-One} & 
    \textbf{One-to-Multi} & 
    \textbf{CAF} & 
    \textbf{CACF} & 
    \textbf{LS} & 
    \textbf{TMC} \\
    \midrule
    MELD & 61.34  & 65.66  & 62.73  & 47.77  & 57.92  & 66.37  & 64.02  & 65.54  & 62.22  & 61.60  & 65.56  \\
    IEMOCAP & 54.96  & 54.59  & 54.82  & 32.64  & 48.14  & 54.49  & 54.16  & 54.26  & 55.06  & 54.30  & 51.22  \\
    MAMI & 70.00  & 66.47  & 66.13  & 67.87  & 66.63  & 68.97  & 64.50  & 65.40  & 67.23  & 67.80  & 69.73  \\
    Memotion & 77.89  & 77.75  & 78.14  & 78.09  & 78.09  & 78.09  & 78.09  & 78.14  & 78.09  & 78.19  & 78.09  \\
    MUTE & 67.87  & 67.31  & 66.59  & 65.63  & 68.19  & 66.03  & 66.99  & 65.87  & 67.07  & 67.95  & 68.67  \\
    MultiOFF & 56.95  & 59.86  & 55.38  & 57.76  & 59.11  & 54.38  & 59.33  & 59.76  & 62.03  & 54.16  & 60.01  \\
    MET-Meme(C) & 35.67  & 34.87  & 36.92  & 29.68  & 24.12  & 33.55  & 35.09  & 32.89  & 36.62  & 34.94  & 33.85  \\
    MET-Meme(E) & 42.95  & 42.47  & 41.83  & 32.69  & 29.81  & 44.23  & 39.74  & 40.38  & 41.03  & 43.11  & 42.47  \\
    Twitter2015 & 76.05  & 76.37  & 75.76  & 63.39  & 68.05  & 74.12  & 70.40  & 75.31  & 75.70  & 75.86  & 64.45  \\
    Twitter1517 & 76.83  & 76.72  & 76.68  & 76.76  & 76.68  & 76.29  & 76.72  & 76.54  & 75.97  & 76.15  & 76.86  \\
     CHSIMS(MSE) & 0.4835  & 0.7431  & 0.4790  & 0.4775  & 0.4804  & 0.4772  & 0.4836  & 0.4794  & 0.4824  & 0.4898 & - \\
    CHSIMS-v2(MSE) & 0.3202  & 0.3566  & 0.3338  & 0.3214  & 0.3391  & 0.3351  & 0.3448  & 0.2801  & 0.3361  & 0.3360  & - \\
    \bottomrule
  \end{tabular}
  }
\end{table*}
\begin{table*}[htbp]
  \centering
  \caption{Performance on Other Datasets.}
  \label{tab:others}
  \resizebox{\textwidth}{!}{%
  \begin{tabular}{@{}l *{11}{>{\centering\arraybackslash}p{1.2cm}} @{}}
    \toprule
    \textbf{Dataset} & 
    \textbf{Concat} & 
    \textbf{TF} & 
    \textbf{Concat Early} & 
    \textbf{LFT} & 
    \textbf{EFT} & 
    \textbf{Multi-to-One} & 
    \textbf{One-to-Multi} & 
    \textbf{CAF} & 
    \textbf{CACF} & 
    \textbf{LS} & 
    \textbf{TMC} \\
    \midrule
    MIRFLICKR & 62.81  & 62.33  & 62.42  & 50.25  & 38.02  & 58.44  & 59.95  & 59.13  & 62.51  & 62.45  & 62.00  \\
    CUB Image-Caption & 77.90 & 76.26 & 78.04 & 2.89 & 1.81 & 71.90  & 69.19 & 21.09 & 73.95 & 79.48 & 77.73 \\
    SUN-RGBD & 60.28 &	59.43 &	60.83 & 45.22  & 31.61  & 53.52  & 56.97 & 53.53  & 58.14 &	60.78 & 59.00  \\
    NYUDv2 & 59.02  & 61.47  & 58.82  & 47.96  & 31.70  & 59.58  & 63.20  & 60.55  & 64.02  & 66.87  & 66.16  \\
    UPMC-Food101 & 91.95 & 92.04 & 91.80 & 83.76 & 8.75 & 86.04 & 88.80 & 86.89 & 90.12 & 91.66 & 92.02 \\
    MVSA-Single & 79.83  & 78.36  & 78.29  & 68.40  & 63.39  & 79.32  & 77.78  & 79.51  & 78.03  & 79.25  & 67.50  \\
    MNIST-SVHN & 96.41 & 96.64 & 96.42 & 62.11 & 93.06 & 95.45 & 93.47 & 95.35 & 96.45 & 96.46 & 96.95 \\
    N-MNIST+N-TIDIGITS & 94.99  & 94.28  & 95.26  & 80.99  & 30.45  & 93.52  & 94.34  & 94.06  & 95.26  & 94.88  & 94.23  \\
    E-MNIST+EEG & 58.72  & 58.21  & 61.28  & 17.69  & 7.95  & 42.56  & 30.51  & 49.74  & 59.49  & 62.05  & 57.69  \\
    \bottomrule
  \end{tabular}
  }
\end{table*}
\section{Experiment and Discussion}

\subsection{Setup}
Using MULTIBENCH++, we load each of the expanded datasets and systematically evaluate the multimodal approaches in our MULTIBENCH++ Algorithms metioned in Sec. \ref{algorithms}: We maintain a consistent experimental setup, varying only the method while keeping all other factors constant, including the training loop and data preprocessing steps. This approach ensures that observed differences in performance can be directly attributed to the fusion method under evaluation. 
We compare our method with several classic baseline approaches previously proposed, including Concat, TensorFusion (\texttt{TF}), ConcatEarly, LateFusionTransformer (\texttt{LFT}), EarlyFusionTransformer (\texttt{EFT}) \citep{liang2021multibench}.

\subsection{Overall performance}
The performance metrics across diverse datasets highlight the nuanced effectiveness of different fusion strategies. As shown in \cref{tab:rs,tab:med,tab:ac,tab:others}, we have the following observations:
(i) Our algorithms (\texttt{CAF, CACF, Logit Summation, TMC, One-to-Multi, Multi-to-One}) yield the highest accuracy on 25 of 37 datasets, routinely beating plain concatenation. (ii) Early fusion like \texttt{Concat} and \texttt{TF} collapses on weakly-aligned modalities, yet shows no gain on saturated tasks (\textbf{SIIM-ISIC}, \textbf{eICU}). (iii) \texttt{CACF} tops seven benchmarks, confirming its broad efficacy.\\
Not surprisingly, the marginal utility of advanced fusion is strictly positive when and only when cross-modal redundancy is low; otherwise, naive concatenation attains near-optimal performance once any single modality approaches the task ceiling. Full results are provided in the appendix.
\subsection{Data Complexity as a Model Selector}
An analysis of the performance metrics in the \cref{tab:rs,tab:med,tab:ac,tab:others} reveals that data complexity is a critical factor for model selection. Taking Table \ref{tab:rs} as an example, on the low-complexity \textbf{Trento} dataset, a simple model like \texttt{Concat} (98.43) is highly effective and performs nearly as well as the top model, \texttt{CACF} (98.68), indicating that increased model complexity provides little benefit. Conversely, for a high-complexity dataset like \textbf{Houston2013}, there's a vast performance gap; simple models fail (\texttt{Concat} at 76.36) while sophisticated models like \texttt{CACF} (83.66) and \texttt{TMC} (81.52) are essential for achieving high accuracy. This proves that the optimal model choice is not universal; it is dictated by the dataset's inherent complexity, requiring simple models for simple data and advanced architectures for complex data.

\section{Future Work and Conclusion}
Multimodal fusion’s future hinges on two challenges: datasets lack fine-grained alignment for real validity, and models remain fragmented and unscalable. The path forward requires creating datasets with deeper structural correspondences while developing unified, theoretically-grounded fusion frameworks. Ultimately, this evolution must extend to the evaluation process itself, shifting from simple tuning towards ethics-aware meta-learning where fairness and robustness are primary objectives.\\
\textbf{In conclusion}, we present MULTIBENCH++, a rigorously-curated, open-source benchmark that unites 30+ datasets across 15+ modalities and 20+ tasks across specialized domains. Coupled with auto-tuned Transformer and hybrid-logit baselines, it gives researchers a fair testbed to compare new fusion models, making results easier to reproduce and closer to real-world use.

\bibliography{aaai2026}

\begin{thebibliography}{68}
\providecommand{\natexlab}[1]{#1}

\bibitem[{Akiba et~al.(2019)Akiba, Sano, Yanase, Ohta, and Koyama}]{akiba2019optuna}
Akiba, T.; Sano, S.; Yanase, T.; Ohta, T.; and Koyama, M. 2019.
\newblock Optuna: A next-generation hyperparameter optimization framework.
\newblock In \emph{Proceedings of the 25th ACM SIGKDD international conference on knowledge discovery \& data mining}, 2623--2631.

\bibitem[{Alain and Bengio(2016)}]{alain2016understanding}
Alain, G.; and Bengio, Y. 2016.
\newblock Understanding intermediate layers using linear classifier probes.
\newblock \emph{arXiv preprint arXiv:1610.01644}.

\bibitem[{Anumula et~al.(2018)Anumula, Neil, Delbruck, and Liu}]{anumula2018feature}
Anumula, J.; Neil, D.; Delbruck, T.; and Liu, S.-C. 2018.
\newblock Feature representations for neuromorphic audio spike streams.
\newblock \emph{Frontiers in neuroscience}, 12: 23.

\bibitem[{Atrey et~al.(2010)Atrey, Hossain, El~Saddik, and Kankanhalli}]{atrey2010multimodal}
Atrey, P.~K.; Hossain, M.~A.; El~Saddik, A.; and Kankanhalli, M.~S. 2010.
\newblock Multimodal fusion for multimedia analysis: a survey.
\newblock \emph{Multimedia systems}, 16(6): 345--379.

\bibitem[{Azam et~al.(2022)Azam, Khan, Salahuddin, Rehman, Khan, Khan, Kadry, and Gandomi}]{azam2022review}
Azam, M.~A.; Khan, K.~B.; Salahuddin, S.; Rehman, E.; Khan, S.~A.; Khan, M.~A.; Kadry, S.; and Gandomi, A.~H. 2022.
\newblock A review on multimodal medical image fusion: Compendious analysis of medical modalities, multimodal databases, fusion techniques and quality metrics.
\newblock \emph{Computers in biology and medicine}, 144: 105253.

\bibitem[{Baltru{\v{s}}aitis, Ahuja, and Morency(2018)}]{baltruvsaitis2018multimodal}
Baltru{\v{s}}aitis, T.; Ahuja, C.; and Morency, L.-P. 2018.
\newblock Multimodal machine learning: A survey and taxonomy.
\newblock \emph{IEEE transactions on pattern analysis and machine intelligence}, 41(2): 423--443.

\bibitem[{Bennett et~al.(2018)Bennett, Buchman, Boyle, Barnes, Wilson, and Schneider}]{bennett2018religious}
Bennett, D.~A.; Buchman, A.~S.; Boyle, P.~A.; Barnes, L.~L.; Wilson, R.~S.; and Schneider, J.~A. 2018.
\newblock Religious orders study and rush memory and aging project.
\newblock \emph{Journal of Alzheimer’s disease}, 64(s1): S161--S189.

\bibitem[{Busso et~al.(2008)Busso, Bulut, Lee, Kazemzadeh, Mower, Kim, Chang, Lee, and Narayanan}]{busso2008iemocap}
Busso, C.; Bulut, M.; Lee, C.-C.; Kazemzadeh, A.; Mower, E.; Kim, S.; Chang, J.~N.; Lee, S.; and Narayanan, S.~S. 2008.
\newblock IEMOCAP: Interactive emotional dyadic motion capture database.
\newblock \emph{Language resources and evaluation}, 42(4): 335--359.

\bibitem[{Caesar et~al.(2020)Caesar, Bankiti, Lang, Vora, Liong, Xu, Krishnan, Pan, Baldan, and Beijbom}]{caesar2020nuscenes}
Caesar, H.; Bankiti, V.; Lang, A.~H.; Vora, S.; Liong, V.~E.; Xu, Q.; Krishnan, A.; Pan, Y.; Baldan, G.; and Beijbom, O. 2020.
\newblock nuscenes: A multimodal dataset for autonomous driving.
\newblock In \emph{Proceedings of the IEEE/CVF conference on computer vision and pattern recognition}, 11621--11631.

\bibitem[{Chen et~al.(2023)Chen, Su, Wu, and Hua}]{chen2023HFIR}
Chen, D.; Su, W.; Wu, P.; and Hua, B. 2023.
\newblock Joint multimodal sentiment analysis based on information relevance.
\newblock \emph{Information Processing \& Management}, 60(2): 103193.

\bibitem[{Cohen et~al.(2017)Cohen, Afshar, Tapson, and Van~Schaik}]{cohen2017emnist}
Cohen, G.; Afshar, S.; Tapson, J.; and Van~Schaik, A. 2017.
\newblock EMNIST: Extending MNIST to handwritten letters.
\newblock In \emph{2017 international joint conference on neural networks (IJCNN)}, 2921--2926. IEEE.

\bibitem[{Debes et~al.(2014)Debes, Merentitis, Heremans, Hahn, Frangiadakis, Van~Kasteren, Liao, Bellens, Pi{\v{z}}urica, Gautama et~al.}]{debes2014hyperspectral}
Debes, C.; Merentitis, A.; Heremans, R.; Hahn, J.; Frangiadakis, N.; Van~Kasteren, T.; Liao, W.; Bellens, R.; Pi{\v{z}}urica, A.; Gautama, S.; et~al. 2014.
\newblock Hyperspectral and LiDAR data fusion: Outcome of the 2013 GRSS data fusion contest.
\newblock \emph{IEEE Journal of Selected Topics in Applied Earth Observations and Remote Sensing}, 7(6): 2405--2418.

\bibitem[{Fersini et~al.(2022)Fersini, Gasparini, Rizzi, Saibene, Chulvi, Rosso, Lees, and Sorensen}]{fersini2022semeval}
Fersini, E.; Gasparini, F.; Rizzi, G.; Saibene, A.; Chulvi, B.; Rosso, P.; Lees, A.; and Sorensen, J. 2022.
\newblock SemEval-2022 Task 5: Multimedia automatic misogyny identification.
\newblock In \emph{Proceedings of the 16th International Workshop on Semantic Evaluation (SemEval-2022)}, 533--549.

\bibitem[{Frome et~al.(2013)Frome, Corrado, Shlens, Bengio, Dean, Ranzato, and Mikolov}]{frome2013devise}
Frome, A.; Corrado, G.~S.; Shlens, J.; Bengio, S.; Dean, J.; Ranzato, M.; and Mikolov, T. 2013.
\newblock Devise: A deep visual-semantic embedding model.
\newblock \emph{Advances in neural information processing systems}, 26.

\bibitem[{Gader et~al.(2013)Gader, Zare, Close, Aitken, and Tuell}]{gader2013muufl}
Gader, P.; Zare, A.; Close, R.; Aitken, J.; and Tuell, G. 2013.
\newblock MUUFL Gulfport hyperspectral and LiDAR airborne data set.
\newblock \emph{Univ. Florida, Gainesville, FL, USA, Tech. Rep. REP-2013-570}.

\bibitem[{Goyal et~al.(2017)Goyal, Khot, Summers-Stay, Batra, and Parikh}]{goyal2017making}
Goyal, Y.; Khot, T.; Summers-Stay, D.; Batra, D.; and Parikh, D. 2017.
\newblock Making the v in vqa matter: Elevating the role of image understanding in visual question answering.
\newblock In \emph{Proceedings of the IEEE conference on computer vision and pattern recognition}, 6904--6913.

\bibitem[{Han et~al.(2022)Han, Zhang, Fu, and Zhou}]{han2022trusted}
Han, Z.; Zhang, C.; Fu, H.; and Zhou, J.~T. 2022.
\newblock Trusted multi-view classification with dynamic evidential fusion.
\newblock \emph{IEEE transactions on pattern analysis and machine intelligence}, 45(2): 2551--2566.

\bibitem[{Hossain, Sharif, and Hoque(2022)}]{hossain2022mute}
Hossain, E.; Sharif, O.; and Hoque, M.~M. 2022.
\newblock MUTE: A multimodal dataset for detecting hateful memes.
\newblock In \emph{Proceedings of the 2nd conference of the asia-pacific chapter of the association for computational linguistics and the 12th international joint conference on natural language processing: student research workshop}, 32--39.

\bibitem[{Hu et~al.(2023)Hu, Liu, Hong, Camero, Yao, Schneider, Kurz, Segl, and Zhu}]{hu2023mdas}
Hu, J.; Liu, R.; Hong, D.; Camero, A.; Yao, J.; Schneider, M.; Kurz, F.; Segl, K.; and Zhu, X.~X. 2023.
\newblock MDAS: A new multimodal benchmark dataset for remote sensing.
\newblock \emph{Earth System Science Data}, 15(1): 113--131.

\bibitem[{Huiskes and Lew(2008)}]{huiskes2008mir}
Huiskes, M.~J.; and Lew, M.~S. 2008.
\newblock The mir flickr retrieval evaluation.
\newblock In \emph{Proceedings of the 1st ACM international conference on Multimedia information retrieval}, 39--43.

\bibitem[{Irvin et~al.(2020)Irvin, Sheng, Ramachandran, Johnson-Yu, Zhou, Story, Rustowicz, Elsworth, Austin, and Ng}]{irvin2020forestnet}
Irvin, J.; Sheng, H.; Ramachandran, N.; Johnson-Yu, S.; Zhou, S.; Story, K.; Rustowicz, R.; Elsworth, C.; Austin, K.; and Ng, A.~Y. 2020.
\newblock Forestnet: Classifying drivers of deforestation in indonesia using deep learning on satellite imagery.
\newblock \emph{arXiv preprint arXiv:2011.05479}.

\bibitem[{Johnson et~al.(2019)Johnson, Pollard, Berkowitz, Greenbaum, Lungren, Deng, Mark, and Horng}]{johnson2019mimic}
Johnson, A.~E.; Pollard, T.~J.; Berkowitz, S.~J.; Greenbaum, N.~R.; Lungren, M.~P.; Deng, C.-y.; Mark, R.~G.; and Horng, S. 2019.
\newblock MIMIC-CXR, a de-identified publicly available database of chest radiographs with free-text reports.
\newblock \emph{Scientific data}, 6(1): 317.

\bibitem[{Johnson et~al.(2016)Johnson, Pollard, Shen, Lehman, Feng, Ghassemi, Moody, Szolovits, Anthony~Celi, and Mark}]{johnson2016mimic}
Johnson, A.~E.; Pollard, T.~J.; Shen, L.; Lehman, L.-w.~H.; Feng, M.; Ghassemi, M.; Moody, B.; Szolovits, P.; Anthony~Celi, L.; and Mark, R.~G. 2016.
\newblock MIMIC-III, a freely accessible critical care database.
\newblock \emph{Scientific data}, 3(1): 1--9.

\bibitem[{Kawahara et~al.(2018)Kawahara, Daneshvar, Argenziano, and Hamarneh}]{kawahara2018seven}
Kawahara, J.; Daneshvar, S.; Argenziano, G.; and Hamarneh, G. 2018.
\newblock Seven-point checklist and skin lesion classification using multitask multimodal neural nets.
\newblock \emph{IEEE journal of biomedical and health informatics}, 23(2): 538--546.

\bibitem[{Kiros, Salakhutdinov, and Zemel(2014)}]{kiros2014unifying}
Kiros, R.; Salakhutdinov, R.; and Zemel, R.~S. 2014.
\newblock Unifying visual-semantic embeddings with multimodal neural language models.
\newblock \emph{arXiv preprint arXiv:1411.2539}.

\bibitem[{Kong et~al.(2011)Kong, Cooper, Wang, Gutman, Gao, Chisolm, Sharma, Pan, Van~Meir, Kurc, Moreno, Saltz, and Brat}]{kong2011integrative}
Kong, J.; Cooper, L. A.~D.; Wang, F.; Gutman, D.~A.; Gao, J.; Chisolm, C.; Sharma, A.; Pan, T.; Van~Meir, E.~G.; Kurc, T.~M.; Moreno, C.~S.; Saltz, J.~H.; and Brat, D.~J. 2011.
\newblock Integrative, Multi-modal Analysis of Glioblastoma Using {TCGA} Molecular Data, Pathology Images and Clinical Outcomes.
\newblock \emph{IEEE Transactions on Biomedical Engineering}, 58(12): 3469--3474.

\bibitem[{Li et~al.(2021)Li, Yang, Ross, and Kanazawa}]{li2021ai}
Li, R.; Yang, S.; Ross, D.~A.; and Kanazawa, A. 2021.
\newblock Ai choreographer: Music conditioned 3d dance generation with aist++.
\newblock In \emph{Proceedings of the IEEE/CVF international conference on computer vision}, 13401--13412.

\bibitem[{Li et~al.(2024)Li, Li, Wang, Jiang, Zhang, Zheng, Wang, Zheng, Huang, Zhou et~al.}]{libenchmarking}
Li, Y.; Li, Y.; Wang, X.; Jiang, Y.; Zhang, Z.; Zheng, X.; Wang, H.; Zheng, H.-T.; Huang, F.; Zhou, J.; et~al. 2024.
\newblock Benchmarking Multimodal Retrieval Augmented Generation with Dynamic VQA Dataset and Self-adaptive Planning Agent.
\newblock In \emph{The Thirteenth International Conference on Learning Representations}.

\bibitem[{Liang et~al.(2021)Liang, Lyu, Fan, Wu, Cheng, Wu, Chen, Wu, Lee, Zhu et~al.}]{liang2021multibench}
Liang, P.~P.; Lyu, Y.; Fan, X.; Wu, Z.; Cheng, Y.; Wu, J.; Chen, L.; Wu, P.; Lee, M.~A.; Zhu, Y.; et~al. 2021.
\newblock Multibench: Multiscale benchmarks for multimodal representation learning.
\newblock \emph{Advances in neural information processing systems}, 2021(DB1): 1.

\bibitem[{Liang, Zadeh, and Morency(2024)}]{liang2024foundations}
Liang, P.~P.; Zadeh, A.; and Morency, L.-P. 2024.
\newblock Foundations \& trends in multimodal machine learning: Principles, challenges, and open questions.
\newblock \emph{ACM Computing Surveys}, 56(10): 1--42.

\bibitem[{Lin et~al.(2020)Lin, Yang, Zhang, Liu, Zhou, and Yang}]{lin2020interbert}
Lin, J.; Yang, A.; Zhang, Y.; Liu, J.; Zhou, J.; and Yang, H. 2020.
\newblock Interbert: Vision-and-language interaction for multi-modal pretraining.
\newblock \emph{arXiv preprint arXiv:2003.13198}.

\bibitem[{Lin et~al.(2025)Lin, Wang, Li, Wang, Shi, He, Zhang, Yu, Zhang, Zhang et~al.}]{lin2025resistive}
Lin, N.; Wang, S.; Li, Y.; Wang, B.; Shi, S.; He, Y.; Zhang, W.; Yu, Y.; Zhang, Y.; Zhang, X.; et~al. 2025.
\newblock Resistive memory-based zero-shot liquid state machine for multimodal event data learning.
\newblock \emph{Nature Computational Science}, 5(1): 37--47.

\bibitem[{Liu et~al.(2022)Liu, Yuan, Mao, Liang, Yang, Qiu, Cheng, Li, Xu, and Gao}]{liu2022make}
Liu, Y.; Yuan, Z.; Mao, H.; Liang, Z.; Yang, W.; Qiu, Y.; Cheng, T.; Li, X.; Xu, H.; and Gao, K. 2022.
\newblock Make acoustic and visual cues matter: Ch-sims v2. 0 dataset and av-mixup consistent module.
\newblock In \emph{Proceedings of the 2022 international conference on multimodal interaction}, 247--258.

\bibitem[{Lu et~al.(2018)Lu, Neves, Carvalho, Zhang, and Ji}]{lu2018visual}
Lu, D.; Neves, L.; Carvalho, V.; Zhang, N.; and Ji, H. 2018.
\newblock Visual attention model for name tagging in multimodal social media.
\newblock In \emph{Proceedings of the 56th Annual Meeting of the Association for Computational Linguistics (Volume 1: Long Papers)}, 1990--1999.

\bibitem[{Lu et~al.(2019)Lu, Batra, Parikh, and Lee}]{lu2019vilbert}
Lu, J.; Batra, D.; Parikh, D.; and Lee, S. 2019.
\newblock Vilbert: Pretraining task-agnostic visiolinguistic representations for vision-and-language tasks.
\newblock \emph{Advances in neural information processing systems}, 32.

\bibitem[{Nagrani et~al.(2021)Nagrani, Yang, Arnab, Jansen, Schmid, and Sun}]{nagrani2021attention}
Nagrani, A.; Yang, S.; Arnab, A.; Jansen, A.; Schmid, C.; and Sun, C. 2021.
\newblock Attention bottlenecks for multimodal fusion.
\newblock \emph{Advances in neural information processing systems}, 34: 14200--14213.

\bibitem[{Niu et~al.(2016)Niu, Zhu, Pang, and El~Saddik}]{niu2016sentiment}
Niu, T.; Zhu, S.; Pang, L.; and El~Saddik, A. 2016.
\newblock Sentiment analysis on multi-view social data.
\newblock In \emph{International conference on multimedia modeling}, 15--27. Springer.

\bibitem[{Okujeni, van~der Linden, and Hostert(2016)}]{okujeni2016berlin}
Okujeni, A.; van~der Linden, S.; and Hostert, P. 2016.
\newblock Berlin-urban-gradient dataset 2009-an enmap preparatory flight campaign.

\bibitem[{Orchard et~al.(2015)Orchard, Jayawant, Cohen, and Thakor}]{orchard2015converting}
Orchard, G.; Jayawant, A.; Cohen, G.~K.; and Thakor, N. 2015.
\newblock Converting static image datasets to spiking neuromorphic datasets using saccades.
\newblock \emph{Frontiers in neuroscience}, 9: 437.

\bibitem[{Pollard et~al.(2018)Pollard, Johnson, Raffa, Celi, Mark, and Badawi}]{pollard2018eicu}
Pollard, T.~J.; Johnson, A.~E.; Raffa, J.~D.; Celi, L.~A.; Mark, R.~G.; and Badawi, O. 2018.
\newblock The eICU Collaborative Research Database, a freely available multi-center database for critical care research.
\newblock \emph{Scientific data}, 5(1): 1--13.

\bibitem[{Poria et~al.(2019)Poria, Hazarika, Majumder, Naik, Cambria, and Mihalcea}]{poria2019meld}
Poria, S.; Hazarika, D.; Majumder, N.; Naik, G.; Cambria, E.; and Mihalcea, R. 2019.
\newblock MELD: A Multimodal Multi-Party Dataset for Emotion Recognition in Conversations.
\newblock In \emph{Proceedings of the 57th Annual Meeting of the Association for Computational Linguistics}, 527--536.

\bibitem[{Ramachandram and Taylor(2017)}]{ramachandram2017deep}
Ramachandram, D.; and Taylor, G.~W. 2017.
\newblock Deep multimodal learning: A survey on recent advances and trends.
\newblock \emph{IEEE signal processing magazine}, 34(6): 96--108.

\bibitem[{Rotemberg et~al.(2021)Rotemberg, Kurtansky, Betz-Stablein, Caffery, Chousakos, Codella, Combalia, Dusza, Guitera, Gutman et~al.}]{rotemberg2021patient}
Rotemberg, V.; Kurtansky, N.; Betz-Stablein, B.; Caffery, L.; Chousakos, E.; Codella, N.; Combalia, M.; Dusza, S.; Guitera, P.; Gutman, D.; et~al. 2021.
\newblock A patient-centric dataset of images and metadata for identifying melanomas using clinical context.
\newblock \emph{Scientific data}, 8(1): 34.

\bibitem[{Sharma et~al.(2020)Sharma, Bhageria, Scott, Pykl, Das, Chakraborty, Pulabaigari, and Gamb{\"a}ck}]{sharma2020semeval}
Sharma, C.; Bhageria, D.; Scott, W.; Pykl, S.; Das, A.; Chakraborty, T.; Pulabaigari, V.; and Gamb{\"a}ck, B. 2020.
\newblock SemEval-2020 Task 8: Memotion Analysis-the Visuo-Lingual Metaphor!
\newblock In \emph{Proceedings of the Fourteenth Workshop on Semantic Evaluation}, 759--773.

\bibitem[{Shi et~al.(2019)Shi, Paige, Torr et~al.}]{shi2019variational}
Shi, Y.; Paige, B.; Torr, P.; et~al. 2019.
\newblock Variational mixture-of-experts autoencoders for multi-modal deep generative models.
\newblock \emph{Advances in neural information processing systems}, 32.

\bibitem[{Silberman et~al.(2012)Silberman, Hoiem, Kohli, and Fergus}]{silberman2012indoor}
Silberman, N.; Hoiem, D.; Kohli, P.; and Fergus, R. 2012.
\newblock Indoor segmentation and support inference from rgbd images.
\newblock In \emph{European conference on computer vision}, 746--760. Springer.

\bibitem[{Soleymani et~al.(2017)Soleymani, Garcia, Jou, Schuller, Chang, and Pantic}]{soleymani2017survey}
Soleymani, M.; Garcia, D.; Jou, B.; Schuller, B.; Chang, S.-F.; and Pantic, M. 2017.
\newblock A survey of multimodal sentiment analysis.
\newblock \emph{Image and Vision Computing}, 65: 3--14.

\bibitem[{Song, Lichtenberg, and Xiao(2015)}]{song2015sun}
Song, S.; Lichtenberg, S.~P.; and Xiao, J. 2015.
\newblock Sun rgb-d: A rgb-d scene understanding benchmark suite.
\newblock In \emph{Proceedings of the IEEE conference on computer vision and pattern recognition}, 567--576.

\bibitem[{Suryawanshi et~al.(2020)Suryawanshi, Chakravarthi, Arcan, and Buitelaar}]{suryawanshi2020multimodal}
Suryawanshi, S.; Chakravarthi, B.~R.; Arcan, M.; and Buitelaar, P. 2020.
\newblock Multimodal meme dataset (MultiOFF) for identifying offensive content in image and text.
\newblock In \emph{Proceedings of the second workshop on trolling, aggression and cyberbullying}, 32--41.

\bibitem[{Tao et~al.(2024)Tao, Huang, Xu, Chen, Feng, and Zhao}]{tao2024probing}
Tao, M.; Huang, Q.; Xu, K.; Chen, L.; Feng, Y.; and Zhao, D. 2024.
\newblock Probing Multimodal Large Language Models for Global and Local Semantic Representations.
\newblock In \emph{Proceedings of the 2024 Joint International Conference on Computational Linguistics, Language Resources and Evaluation (LREC-COLING 2024)}, 13050--13056.

\bibitem[{Tsai et~al.(2019)Tsai, Bai, Liang, Kolter, Morency, and Salakhutdinov}]{tsai2019multimodal}
Tsai, Y.-H.~H.; Bai, S.; Liang, P.~P.; Kolter, J.~Z.; Morency, L.-P.; and Salakhutdinov, R. 2019.
\newblock Multimodal transformer for unaligned multimodal language sequences.
\newblock In \emph{Proceedings of the conference. Association for computational linguistics. Meeting}, volume 2019, 6558.

\bibitem[{Wang et~al.(2015)Wang, Kumar, Thome, Cord, and Precioso}]{wang2015recipe}
Wang, X.; Kumar, D.; Thome, N.; Cord, M.; and Precioso, F. 2015.
\newblock Recipe recognition with large multimodal food dataset.
\newblock In \emph{2015 IEEE International Conference on Multimedia \& Expo Workshops (ICMEW)}, 1--6. IEEE.

\bibitem[{Wang et~al.(2022)Wang, Chen, Cao, Huang, Sun, and Wang}]{wang2022multimodal}
Wang, Y.; Chen, X.; Cao, L.; Huang, W.; Sun, F.; and Wang, Y. 2022.
\newblock Multimodal Token Fusion for Vision Transformers.
\newblock In \emph{Proceedings of the {IEEE/CVF} Conference on Computer Vision and Pattern Recognition ({CVPR})}.

\bibitem[{Wei et~al.(2020)Wei, Zhang, Li, Zhang, and Wu}]{wei2020multi}
Wei, X.; Zhang, T.; Li, Y.; Zhang, Y.; and Wu, F. 2020.
\newblock Multi-Modality Cross Attention Network for Image and Sentence Matching.
\newblock In \emph{Proceedings of the {IEEE/CVF} Conference on Computer Vision and Pattern Recognition ({CVPR})}.

\bibitem[{Weinstein et~al.(2013)Weinstein, Collisson, Mills, Shaw, Ozenberger, Ellrott, Shmulevich, Sander, and Stuart}]{weinstein2013cancer}
Weinstein, J.~N.; Collisson, E.~A.; Mills, G.~B.; Shaw, K.~R.; Ozenberger, B.~A.; Ellrott, K.; Shmulevich, I.; Sander, C.; and Stuart, J.~M. 2013.
\newblock The cancer genome atlas pan-cancer analysis project.
\newblock \emph{Nature genetics}, 45(10): 1113--1120.

\bibitem[{Willett et~al.(2021)Willett, Avansino, Hochberg, Henderson, and Shenoy}]{willett2021high}
Willett, F.~R.; Avansino, D.~T.; Hochberg, L.~R.; Henderson, J.~M.; and Shenoy, K.~V. 2021.
\newblock High-performance brain-to-text communication via handwriting.
\newblock \emph{Nature}, 593(7858): 249--254.

\bibitem[{Wu et~al.(2023)Wu, Fang, Li, Fu, Lin, Li, Huang, Yu, Song, Xu et~al.}]{wu2023gamma}
Wu, J.; Fang, H.; Li, F.; Fu, H.; Lin, F.; Li, J.; Huang, Y.; Yu, Q.; Song, S.; Xu, X.; et~al. 2023.
\newblock Gamma challenge: glaucoma grading from multi-modality images.
\newblock \emph{Medical Image Analysis}, 90: 102938.

\bibitem[{Xu et~al.(2022)Xu, Li, Zheng, Naseriparsa, Zhao, Lin, and Xia}]{xu2022met}
Xu, B.; Li, T.; Zheng, J.; Naseriparsa, M.; Zhao, Z.; Lin, H.; and Xia, F. 2022.
\newblock Met-meme: A multimodal meme dataset rich in metaphors.
\newblock In \emph{Proceedings of the 45th international ACM SIGIR conference on research and development in information retrieval}, 2887--2899.

\bibitem[{Xu, Zhu, and Clifton(2023)}]{xu2023multimodal}
Xu, P.; Zhu, X.; and Clifton, D.~A. 2023.
\newblock Multimodal learning with transformers: A survey.
\newblock \emph{IEEE Transactions on Pattern Analysis and Machine Intelligence}, 45(10): 12113--12132.

\bibitem[{Xu et~al.(2018)Xu, Du, Zhang, and Zhang}]{xu2018hyperspectral}
Xu, Y.; Du, B.; Zhang, F.; and Zhang, L. 2018.
\newblock Hyperspectral image classification via a random patches network.
\newblock \emph{ISPRS journal of photogrammetry and remote sensing}, 142: 344--357.

\bibitem[{Yu et~al.(2020)Yu, Xu, Meng, Zhu, Ma, Wu, Zou, and Yang}]{yu2020ch}
Yu, W.; Xu, H.; Meng, F.; Zhu, Y.; Ma, Y.; Wu, J.; Zou, J.; and Yang, K. 2020.
\newblock Ch-sims: A chinese multimodal sentiment analysis dataset with fine-grained annotation of modality.
\newblock In \emph{Proceedings of the 58th annual meeting of the association for computational linguistics}, 3718--3727.

\bibitem[{Zadeh et~al.(2016)Zadeh, Zellers, Pincus, and Morency}]{zadeh2016mosi}
Zadeh, A.; Zellers, R.; Pincus, E.; and Morency, L.-P. 2016.
\newblock Mosi: multimodal corpus of sentiment intensity and subjectivity analysis in online opinion videos.
\newblock \emph{arXiv preprint arXiv:1606.06259}.

\bibitem[{Zadeh et~al.(2018)Zadeh, Liang, Poria, Cambria, and Morency}]{zadeh2018multimodal}
Zadeh, A.~B.; Liang, P.~P.; Poria, S.; Cambria, E.; and Morency, L.-P. 2018.
\newblock Multimodal language analysis in the wild: Cmu-mosei dataset and interpretable dynamic fusion graph.
\newblock In \emph{Proceedings of the 56th Annual Meeting of the Association for Computational Linguistics (Volume 1: Long Papers)}, 2236--2246.

\bibitem[{Zhan et~al.(2021)Zhan, Wu, Dong, Wei, Lu, Zhang, Xu, and Liang}]{zhan2021product1m}
Zhan, X.; Wu, Y.; Dong, X.; Wei, Y.; Lu, M.; Zhang, Y.; Xu, H.; and Liang, X. 2021.
\newblock Product1m: Towards weakly supervised instance-level product retrieval via cross-modal pretraining.
\newblock In \emph{Proceedings of the IEEE/CVF international conference on computer vision}, 11782--11791.

\bibitem[{Zhang et~al.(2018)Zhang, Fu, Liu, and Huang}]{zhang2018adaptive}
Zhang, Q.; Fu, J.; Liu, X.; and Huang, X. 2018.
\newblock Adaptive co-attention network for named entity recognition in tweets.
\newblock In \emph{Proceedings of the AAAI conference on artificial intelligence}, volume~32.

\bibitem[{Zhang et~al.(2024)Zhang, Wei, Han, Fu, Peng, Deng, Hu, Xu, Wen, Hu et~al.}]{zhang2024multimodal}
Zhang, Q.; Wei, Y.; Han, Z.; Fu, H.; Peng, X.; Deng, C.; Hu, Q.; Xu, C.; Wen, J.; Hu, D.; et~al. 2024.
\newblock Multimodal fusion on low-quality data: A comprehensive survey.
\newblock \emph{arXiv preprint arXiv:2404.18947}.

\bibitem[{Zhang et~al.(2023)Zhang, Wu, Zhang, Hu, Fu, Zhou, and Peng}]{zhang2023provable}
Zhang, Q.; Wu, H.; Zhang, C.; Hu, Q.; Fu, H.; Zhou, J.~T.; and Peng, X. 2023.
\newblock Provable dynamic fusion for low-quality multimodal data.
\newblock In \emph{International conference on machine learning}, 41753--41769. PMLR.

\bibitem[{Zhu et~al.(2025)Zhu, Zhou, Qian, He, Zhao, Shah, and Koutra}]{zhu2025mosaic}
Zhu, J.; Zhou, Y.; Qian, S.; He, Z.; Zhao, T.; Shah, N.; and Koutra, D. 2025.
\newblock Mosaic of modalities: A comprehensive benchmark for multimodal graph learning.
\newblock In \emph{Proceedings of the Computer Vision and Pattern Recognition Conference}, 14215--14224.

\end{thebibliography}
\clearpage 
\appendix 

\begin{table*}[htbp]
  \centering
  
  \caption{Performance on Affective Computing \& Social Media Understanding Datasets. 
  Results are shown as \textit{mean} $\pm$ \textit{standard deviation}. 
  For MSE metrics, lower is better. The dash ``-" indicates the method is not applicable.} 
  \label{tab:ac_append}
  \resizebox{\textwidth}{!}{

   \begin{tabular}{@{}l *{11}{c} @{}}
    \toprule
    \textbf{Dataset} & 
    \textbf{Concat} & \textbf{TF} & \textbf{ConcatEarly} & \textbf{LFT} & \textbf{EFT} & 
    \textbf{Multi-to-One} & \textbf{One-to-Multi} & \textbf{CAF} & \textbf{CACF} & \textbf{LS} & \textbf{TMC} \\
    \midrule
    MELD & \makecell{61.34 \\ {\scriptsize $\pm$0.55}} & \makecell{65.66 \\ {\scriptsize $\pm$0.71}} & \makecell{62.73 \\ {\scriptsize $\pm$0.05}} & \makecell{47.77 \\ {\scriptsize $\pm$0.57}} & \makecell{57.92 \\ {\scriptsize $\pm$0.82}} & \makecell{66.37 \\ {\scriptsize $\pm$0.78}} & \makecell{64.02 \\ {\scriptsize $\pm$0.40}} & \makecell{65.54 \\ {\scriptsize $\pm$0.66}} & \makecell{62.22 \\ {\scriptsize $\pm$0.58}} & \makecell{61.60 \\ {\scriptsize $\pm$0.12}} & \makecell{65.56 \\ {\scriptsize $\pm$0.18}} \\
    IEMOCAP & \makecell{54.96 \\ {\scriptsize $\pm$0.13}} & \makecell{54.59 \\ {\scriptsize $\pm$0.13}} & \makecell{54.82 \\ {\scriptsize $\pm$0.08}} & \makecell{32.64 \\ {\scriptsize $\pm$0.59}} & \makecell{48.14 \\ {\scriptsize $\pm$1.93}} & \makecell{54.49 \\ {\scriptsize $\pm$0.35}} & \makecell{54.16 \\ {\scriptsize $\pm$0.74}} & \makecell{54.26 \\ {\scriptsize $\pm$0.19}} & \makecell{55.06 \\ {\scriptsize $\pm$0.32}} & \makecell{54.30 \\ {\scriptsize $\pm$0.24}} & \makecell{51.22 \\ {\scriptsize $\pm$0.15}} \\
    MAMI & \makecell{70.00 \\ {\scriptsize $\pm$1.00}} & \makecell{66.47 \\ {\scriptsize $\pm$3.84}} & \makecell{66.13 \\ {\scriptsize $\pm$5.43}} & \makecell{67.87 \\ {\scriptsize $\pm$2.49}} & \makecell{66.63 \\ {\scriptsize $\pm$2.28}} & \makecell{68.97 \\ {\scriptsize $\pm$2.53}} & \makecell{64.50 \\ {\scriptsize $\pm$3.69}} & \makecell{65.40 \\ {\scriptsize $\pm$3.06}} & \makecell{67.23 \\ {\scriptsize $\pm$1.90}} & \makecell{67.80 \\ {\scriptsize $\pm$2.61}} & \makecell{69.73 \\ {\scriptsize $\pm$1.03}} \\
    Memotion & \makecell{77.89 \\ {\scriptsize $\pm$2.51}} & \makecell{77.75 \\ {\scriptsize $\pm$2.02}} & \makecell{78.14 \\ {\scriptsize $\pm$4.39}} & \makecell{78.09 \\ {\scriptsize $\pm$2.14}} & \makecell{78.09 \\ {\scriptsize $\pm$1.71}} & \makecell{78.09 \\ {\scriptsize $\pm$5.01}} & \makecell{78.09 \\ {\scriptsize $\pm$1.71}} & \makecell{78.14 \\ {\scriptsize $\pm$5.00}} & \makecell{78.09 \\ {\scriptsize $\pm$1.80}} & \makecell{78.19 \\ {\scriptsize $\pm$2.61}} & \makecell{78.09 \\ {\scriptsize $\pm$1.94}} \\
    MUTE & \makecell{67.87 \\ {\scriptsize $\pm$1.22}} & \makecell{67.31 \\ {\scriptsize $\pm$1.85}} & \makecell{66.59 \\ {\scriptsize $\pm$1.52}} & \makecell{65.63 \\ {\scriptsize $\pm$3.74}} & \makecell{68.19 \\ {\scriptsize $\pm$2.47}} & \makecell{66.03 \\ {\scriptsize $\pm$3.35}} & \makecell{66.99 \\ {\scriptsize $\pm$2.16}} & \makecell{65.87 \\ {\scriptsize $\pm$3.34}} & \makecell{67.07 \\ {\scriptsize $\pm$2.52}} & \makecell{67.95 \\ {\scriptsize $\pm$0.89}} & \makecell{68.67 \\ {\scriptsize $\pm$3.24}} \\
    MultiOFF & \makecell{56.95 \\ {\scriptsize $\pm$1.46}} & \makecell{59.86 \\ {\scriptsize $\pm$1.22}} & \makecell{55.38 \\ {\scriptsize $\pm$2.00}} & \makecell{57.76 \\ {\scriptsize $\pm$5.43}} & \makecell{59.11 \\ {\scriptsize $\pm$6.82}} & \makecell{54.38 \\ {\scriptsize $\pm$2.41}} & \makecell{59.33 \\ {\scriptsize $\pm$2.49}} & \makecell{59.76 \\ {\scriptsize $\pm$1.87}} & \makecell{62.03 \\ {\scriptsize $\pm$0.86}} & \makecell{54.16 \\ {\scriptsize $\pm$1.10}} & \makecell{60.01 \\ {\scriptsize $\pm$8.02}} \\
    MET-Meme(C) & \makecell{35.67 \\ {\scriptsize $\pm$0.00}} & \makecell{34.87 \\ {\scriptsize $\pm$0.00}} & \makecell{36.92 \\ {\scriptsize $\pm$0.00}} & \makecell{29.68 \\ {\scriptsize $\pm$0.41}} & \makecell{24.12 \\ {\scriptsize $\pm$0.45}} & \makecell{33.55 \\ {\scriptsize $\pm$0.10}} & \makecell{35.09 \\ {\scriptsize $\pm$0.43}} & \makecell{32.89 \\ {\scriptsize $\pm$0.05}} & \makecell{36.62 \\ {\scriptsize $\pm$0.63}} & \makecell{34.94 \\ {\scriptsize $\pm$0.68}} & \makecell{33.85 \\ {\scriptsize $\pm$0.33}} \\
    MET-Meme(E) & \makecell{42.95 \\ {\scriptsize $\pm$2.23}} & \makecell{42.47 \\ {\scriptsize $\pm$3.87}} & \makecell{41.83 \\ {\scriptsize $\pm$2.04}} & \makecell{32.69 \\ {\scriptsize $\pm$1.71}} & \makecell{29.81 \\ {\scriptsize $\pm$2.67}} & \makecell{44.23 \\ {\scriptsize $\pm$4.21}} & \makecell{39.74 \\ {\scriptsize $\pm$4.95}} & \makecell{40.38 \\ {\scriptsize $\pm$1.56}} & \makecell{41.03 \\ {\scriptsize $\pm$2.61}} & \makecell{43.11 \\ {\scriptsize $\pm$1.94}} & \makecell{42.47 \\ {\scriptsize $\pm$3.00}} \\
    Twitter2015 & \makecell{76.05 \\ {\scriptsize $\pm$0.36}} & \makecell{76.37 \\ {\scriptsize $\pm$0.28}} & \makecell{75.76 \\ {\scriptsize $\pm$0.09}} & \makecell{63.39 \\ {\scriptsize $\pm$0.41}} & \makecell{68.05 \\ {\scriptsize $\pm$0.45}} & \makecell{74.12 \\ {\scriptsize $\pm$0.10}} & \makecell{70.40 \\ {\scriptsize $\pm$0.43}} & \makecell{75.31 \\ {\scriptsize $\pm$0.05}} & \makecell{75.70 \\ {\scriptsize $\pm$0.63}} & \makecell{75.86 \\ {\scriptsize $\pm$0.68}} & \makecell{64.45 \\ {\scriptsize $\pm$0.33}} \\
    Twitter2017 & \makecell{76.83 \\ {\scriptsize $\pm$1.34}} & \makecell{76.72 \\ {\scriptsize $\pm$1.09}} & \makecell{76.68 \\ {\scriptsize $\pm$0.52}} & \makecell{76.76 \\ {\scriptsize $\pm$1.19}} & \makecell{76.68 \\ {\scriptsize $\pm$1.82}} & \makecell{76.29 \\ {\scriptsize $\pm$1.01}} & \makecell{76.72 \\ {\scriptsize $\pm$2.49}} & \makecell{76.54 \\ {\scriptsize $\pm$1.87}} & \makecell{75.97 \\ {\scriptsize $\pm$1.47}} & \makecell{76.15 \\ {\scriptsize $\pm$2.34}} & \makecell{76.86 \\ {\scriptsize $\pm$0.41}} \\
    \midrule
    \multicolumn{12}{c}{\textit{Lower is better for MSE metrics}} \\
    \cmidrule(l){1-12}
    \textbf{Dataset (MSE)} &  \textbf{Concat} & \textbf{TF} & \textbf{ConcatEarly} & \textbf{LFT} & \textbf{EFT} & 
    \textbf{Multi-to-One} & \textbf{One-to-Multi} & \textbf{CAF} & \textbf{CACF} & \textbf{LS} & \textbf{TMC} \\
    \midrule
    CHSIMS & \makecell{0.4835 \\ {\scriptsize $\pm$0.0083}} & \makecell{0.7431 \\ {\scriptsize $\pm$0.3012}} & \makecell{0.4790 \\ {\scriptsize $\pm$0.0027}} & \makecell{0.4775 \\ {\scriptsize $\pm$0.0019}} & \makecell{0.4804 \\ {\scriptsize $\pm$0.0021}} & \makecell{0.4772 \\ {\scriptsize $\pm$0.0034}} & \makecell{0.4836 \\ {\scriptsize $\pm$0.0060}} & \makecell{0.4794 \\ {\scriptsize $\pm$0.0042}} & \makecell{0.4824 \\ {\scriptsize $\pm$0.0030}} & \makecell{0.4898 \\ {\scriptsize $\pm$0.0031}} & - \\
    CHSIMS-v2 & \makecell{0.3202 \\ {\scriptsize $\pm$0.0301}} & \makecell{0.3566 \\ {\scriptsize $\pm$0.0246}} & \makecell{0.3338 \\ {\scriptsize $\pm$0.0004}} & \makecell{0.3214 \\ {\scriptsize $\pm$0.0224}} & \makecell{0.3391 \\ {\scriptsize $\pm$0.0068}} & \makecell{0.3351 \\ {\scriptsize $\pm$0.0028}} & \makecell{0.3448 \\ {\scriptsize $\pm$0.0067}} & \makecell{0.2801 \\ {\scriptsize $\pm$0.0381}} & \makecell{0.3361 \\ {\scriptsize $\pm$0.0009}} & \makecell{0.3360 \\ {\scriptsize $\pm$0.0024}} & - \\
    \bottomrule
  \end{tabular}
  }
\end{table*}

\begin{strip}
  \centering
  \LARGE
  \bfseries Appendix
\end{strip}

\section{More Exprimental Results}
This section contains the complete set of quantitative results that support the main paper.  Every reported metric is the mean of 3 independent training runs initialized with different random seeds.  After each run we computed the metric on the test split and the population standard deviation  across the 3 runs is reported in parentheses immediately after the mean. Due to their large scale, the results for MNIST-SVHN, UPMC-Food101, and CUB Image-Caption are not reported after only a single run.
\begin{table*}[htbp]
  \centering
  \caption{Performance on Medical AI Datasets. 
  Results are shown as \textit{mean} with the $\pm$\,\textit{standard deviation} value below it in a smaller font.
  The dash ``-" indicates the method is not applicable. 
 }
  \label{tab:med_append}

  \renewcommand{\cellalign}{c} 
  \resizebox{\textwidth}{!}{
  \begin{tabular}{@{}l *{11}{c} @{}}
    \toprule
   \textbf{Dataset} & 
    \textbf{Concat} & \textbf{TF} & \textbf{ConcatEarly} & \textbf{LFT} & \textbf{EFT} & 
    \textbf{Multi-to-One} & \textbf{One-to-Multi} & \textbf{CAF} & \textbf{CACF} & \textbf{LS} & \textbf{TMC} \\
    \midrule
    TCGA-BRCA & \makecell{78.17 \\ {\scriptsize $\pm$1.71}} & \makecell{77.18 \\ {\scriptsize $\pm$2.85}} & \makecell{77.18 \\ {\scriptsize $\pm$0.74}} & \makecell{69.44 \\ {\scriptsize $\pm$7.87}} & \makecell{65.67 \\ {\scriptsize $\pm$4.87}} & \makecell{76.79 \\ {\scriptsize $\pm$1.68}} & \makecell{76.19 \\ {\scriptsize $\pm$3.67}} & \makecell{78.77 \\ {\scriptsize $\pm$0.28}} & \makecell{78.37 \\ {\scriptsize $\pm$2.02}} & \makecell{77.18 \\ {\scriptsize $\pm$1.40}} & \makecell{75.40 \\ {\scriptsize $\pm$0.56}} \\
    ROSMAP & \makecell{70.95 \\ {\scriptsize $\pm$3.09}} & \makecell{75.71 \\ {\scriptsize $\pm$0.67}} & \makecell{70.00 \\ {\scriptsize $\pm$2.33}} & \makecell{42.86 \\ {\scriptsize $\pm$0.67}} & \makecell{69.52 \\ {\scriptsize $\pm$1.35}} & \makecell{69.52 \\ {\scriptsize $\pm$0.67}} & \makecell{68.10 \\ {\scriptsize $\pm$0.67}} & \makecell{66.67 \\ {\scriptsize $\pm$3.09}} & \makecell{71.90 \\ {\scriptsize $\pm$0.09}} & \makecell{71.43 \\ {\scriptsize $\pm$3.02}} & \makecell{66.67 \\ {\scriptsize $\pm$5.99}} \\
    SIM-ISIC & \makecell{97.86 \\ {\scriptsize $\pm$0.07}} & \makecell{97.77 \\ {\scriptsize $\pm$0.04}} & \makecell{97.86 \\ {\scriptsize $\pm$0.07}} & \makecell{97.83 \\ {\scriptsize $\pm$0.00}} & \makecell{97.85 \\ {\scriptsize $\pm$0.07}} & \makecell{97.85 \\ {\scriptsize $\pm$0.03}} & \makecell{97.83 \\ {\scriptsize $\pm$0.33}} & \makecell{97.86 \\ {\scriptsize $\pm$0.07}} & \makecell{97.83 \\ {\scriptsize $\pm$0.00}} & \makecell{97.83 \\ {\scriptsize $\pm$0.02}} & \makecell{97.85 \\ {\scriptsize $\pm$0.20}} \\
    Derm7pt & \makecell{45.49 \\ {\scriptsize $\pm$2.25}} & \makecell{52.72 \\ {\scriptsize $\pm$0.84}} & \makecell{45.41 \\ {\scriptsize $\pm$1.44}} & \makecell{43.96 \\ {\scriptsize $\pm$0.32}} & \makecell{38.86 \\ {\scriptsize $\pm$1.11}} & \makecell{45.92 \\ {\scriptsize $\pm$0.60}} & \makecell{46.77 \\ {\scriptsize $\pm$2.19}} & \makecell{48.13 \\ {\scriptsize $\pm$1.36}} & \makecell{52.72 \\ {\scriptsize $\pm$2.30}} & \makecell{46.34 \\ {\scriptsize $\pm$1.22}} & \makecell{52.72 \\ {\scriptsize $\pm$1.22}} \\
    GAMMA & \makecell{61.43 \\ {\scriptsize $\pm$2.33}} & \makecell{62.86 \\ {\scriptsize $\pm$3.50}} & \makecell{63.81 \\ {\scriptsize $\pm$0.67}} & \makecell{62.38 \\ {\scriptsize $\pm$2.43}} & \makecell{59.05 \\ {\scriptsize $\pm$0.67}} & \makecell{57.62 \\ {\scriptsize $\pm$2.94}} & \makecell{65.71 \\ {\scriptsize $\pm$4.67}} & \makecell{63.33 \\ {\scriptsize $\pm$2.94}} & \makecell{63.33 \\ {\scriptsize $\pm$1.35}} & \makecell{62.38 \\ {\scriptsize $\pm$2.33}} & \makecell{61.43 \\ {\scriptsize $\pm$2.33}} \\
    MIMIC-III & \makecell{68.09 \\ {\scriptsize $\pm$0.91}} & \makecell{68.81 \\ {\scriptsize $\pm$0.23}} & \makecell{68.48 \\ {\scriptsize $\pm$0.73}} & \makecell{68.42 \\ {\scriptsize $\pm$0.15}} & \makecell{68.59 \\ {\scriptsize $\pm$0.11}} & \makecell{69.01 \\ {\scriptsize $\pm$0.33}} & \makecell{68.76 \\ {\scriptsize $\pm$0.35}} & \makecell{68.51 \\ {\scriptsize $\pm$0.25}} & \makecell{68.88 \\ {\scriptsize $\pm$0.27}} & \makecell{68.47 \\ {\scriptsize $\pm$0.22}} & \makecell{68.87 \\ {\scriptsize $\pm$0.20}} \\
    eICU & \makecell{90.05 \\ {\scriptsize $\pm$0.00}} & \makecell{90.03 \\ {\scriptsize $\pm$0.03}} & \makecell{90.05 \\ {\scriptsize $\pm$0.00}} & \makecell{90.05 \\ {\scriptsize $\pm$0.00}} & \makecell{90.05 \\ {\scriptsize $\pm$0.00}} & \makecell{90.05 \\ {\scriptsize $\pm$0.00}} & \makecell{90.05 \\ {\scriptsize $\pm$0.03}} & \makecell{90.07 \\ {\scriptsize $\pm$0.00}} & \makecell{90.05 \\ {\scriptsize $\pm$0.00}} & \makecell{90.05 \\ {\scriptsize $\pm$0.03}} & \makecell{90.03 \\ {\scriptsize $\pm$0.03}} \\
    TCGA & \makecell{51.91 \\ {\scriptsize $\pm$4.09}} & - & \makecell{53.55 \\ {\scriptsize $\pm$4.09}} & \makecell{60.66 \\ {\scriptsize $\pm$5.35}} & \makecell{51.37 \\ {\scriptsize $\pm$2.79}} & \makecell{60.11 \\ {\scriptsize $\pm$3.86}} & \makecell{53.01 \\ {\scriptsize $\pm$6.04}} & \makecell{56.28 \\ {\scriptsize $\pm$5.41}} & \makecell{61.75 \\ {\scriptsize $\pm$2.04}} & \makecell{54.10 \\ {\scriptsize $\pm$6.13}} & \makecell{62.84 \\ {\scriptsize $\pm$6.18}} \\
    \midrule
    \multicolumn{12}{c}{\textit{For MIMIC-CXR, the metric is macro-F1}} \\
    \cmidrule(l){1-12}
    MIMIC-CXR & \makecell{0.6861 \\ {\scriptsize $\pm$0.0248}} & \makecell{0.8458 \\ {\scriptsize $\pm$0.0447}} & \makecell{0.6788 \\ {\scriptsize $\pm$0.0527}} & \makecell{0.1551 \\ {\scriptsize $\pm$0.0321}} & \makecell{0.1829 \\ {\scriptsize $\pm$0.0131}} & \makecell{0.5229 \\ {\scriptsize $\pm$0.1516}} & \makecell{0.4031 \\ {\scriptsize $\pm$0.0852}} & \makecell{0.6136 \\ {\scriptsize $\pm$0.0076}} & \makecell{0.7523 \\ {\scriptsize $\pm$0.0042}} & \makecell{0.7644 \\ {\scriptsize $\pm$0.0171}} & - \\
    \bottomrule
  \end{tabular}
  }
\end{table*}
\begin{table*}[htbp]
  \centering
  \caption{Performance on Remote Sensing Datasets. 
  Results are shown as \textit{mean} with the $\pm$\,\textit{standard deviation} value below it in a smaller font.
  The dash ``-" indicates the method is not applicable. 
 }
  \label{tab:rs_append}
  
  \small 
  \renewcommand{\cellalign}{c} 
\resizebox{\textwidth}{!}{
  \begin{tabular}{@{}l *{11}{c} @{}}
    \toprule
    \textbf{Dataset} & 
    \textbf{Concat} & \textbf{TF} & \textbf{ConcatEarly} & \textbf{LFT} & \textbf{EFT} & 
    \textbf{Multi-to-One} & \textbf{One-to-Multi} & \textbf{CAF} & \textbf{CACF} & \textbf{LS} & \textbf{TMC} \\
    \midrule
    Houston2013 & \makecell{76.36 \\ {\scriptsize $\pm$1.19}} & \makecell{79.72 \\ {\scriptsize $\pm$0.29}} & \makecell{79.59 \\ {\scriptsize $\pm$0.31}} & \makecell{69.27 \\ {\scriptsize $\pm$4.50}} & \makecell{21.60 \\ {\scriptsize $\pm$1.82}} & \makecell{78.56 \\ {\scriptsize $\pm$0.63}} & \makecell{79.97 \\ {\scriptsize $\pm$0.41}} & \makecell{74.12 \\ {\scriptsize $\pm$1.70}} & \makecell{83.66 \\ {\scriptsize $\pm$0.13}} & \makecell{80.55 \\ {\scriptsize $\pm$0.47}} & \makecell{81.52 \\ {\scriptsize $\pm$0.55}} \\
    Houston2018 & \makecell{79.30 \\ {\scriptsize $\pm$1.14}} & \makecell{79.24 \\ {\scriptsize $\pm$1.15}} & \makecell{78.52 \\ {\scriptsize $\pm$2.24}} & \makecell{66.30 \\ {\scriptsize $\pm$1.04}} & \makecell{48.58 \\ {\scriptsize $\pm$3.92}} & \makecell{71.34 \\ {\scriptsize $\pm$2.06}} & \makecell{74.52 \\ {\scriptsize $\pm$1.29}} & \makecell{75.99 \\ {\scriptsize $\pm$1.73}} & \makecell{82.20 \\ {\scriptsize $\pm$2.20}} & \makecell{78.09 \\ {\scriptsize $\pm$1.21}} & \makecell{78.52 \\ {\scriptsize $\pm$0.89}} \\
    Berlin & \makecell{79.03 \\ {\scriptsize $\pm$1.00}} & \makecell{74.45 \\ {\scriptsize $\pm$1.91}} & \makecell{79.47 \\ {\scriptsize $\pm$0.59}} & \makecell{64.23 \\ {\scriptsize $\pm$3.80}} & \makecell{62.63 \\ {\scriptsize $\pm$3.93}} & \makecell{69.95 \\ {\scriptsize $\pm$6.48}} & \makecell{68.75 \\ {\scriptsize $\pm$6.91}} & \makecell{75.01 \\ {\scriptsize $\pm$0.14}} & \makecell{78.26 \\ {\scriptsize $\pm$0.22}} & \makecell{78.91 \\ {\scriptsize $\pm$0.83}} & \makecell{77.93 \\ {\scriptsize $\pm$0.83}} \\
    MUUFL Gulfport & \makecell{84.21 \\ {\scriptsize $\pm$2.92}} & \makecell{83.90 \\ {\scriptsize $\pm$1.65}} & \makecell{86.26 \\ {\scriptsize $\pm$1.13}} & \makecell{71.77 \\ {\scriptsize $\pm$3.43}} & \makecell{46.68 \\ {\scriptsize $\pm$3.16}} & \makecell{80.95 \\ {\scriptsize $\pm$1.10}} & \makecell{81.23 \\ {\scriptsize $\pm$2.41}} & \makecell{83.99 \\ {\scriptsize $\pm$1.02}} & \makecell{86.42 \\ {\scriptsize $\pm$0.67}} & \makecell{86.64 \\ {\scriptsize $\pm$0.64}} & \makecell{83.20 \\ {\scriptsize $\pm$4.32}} \\
    Trento & \makecell{98.43 \\ {\scriptsize $\pm$0.26}} & \makecell{96.95 \\ {\scriptsize $\pm$1.32}} & \makecell{98.14 \\ {\scriptsize $\pm$0.63}} & \makecell{95.68 \\ {\scriptsize $\pm$0.84}} & \makecell{71.64 \\ {\scriptsize $\pm$4.56}} & \makecell{97.71 \\ {\scriptsize $\pm$0.10}} & \makecell{96.08 \\ {\scriptsize $\pm$1.39}} & \makecell{97.74 \\ {\scriptsize $\pm$0.65}} & \makecell{98.68 \\ {\scriptsize $\pm$0.20}} & \makecell{98.53 \\ {\scriptsize $\pm$0.36}} & \makecell{97.67 \\ {\scriptsize $\pm$1.02}} \\
    Augsburg & \makecell{89.24 \\ {\scriptsize $\pm$0.86}} & \makecell{85.86 \\ {\scriptsize $\pm$1.03}} & \makecell{89.50 \\ {\scriptsize $\pm$1.13}} & \makecell{82.88 \\ {\scriptsize $\pm$0.50}} & \makecell{57.29 \\ {\scriptsize $\pm$5.40}} & \makecell{85.86 \\ {\scriptsize $\pm$0.73}} & \makecell{86.80 \\ {\scriptsize $\pm$0.91}} & \makecell{87.92 \\ {\scriptsize $\pm$0.34}} & \makecell{89.05 \\ {\scriptsize $\pm$0.35}} & \makecell{89.49 \\ {\scriptsize $\pm$0.34}} & \makecell{87.21 \\ {\scriptsize $\pm$0.63}} \\
    ForestNet & \makecell{45.18 \\ {\scriptsize $\pm$2.14}} & \makecell{45.63 \\ {\scriptsize $\pm$1.23}} & \makecell{45.68 \\ {\scriptsize $\pm$0.38}} & \makecell{47.19 \\ {\scriptsize $\pm$1.99}} & \makecell{44.33 \\ {\scriptsize $\pm$0.93}} & \makecell{45.78 \\ {\scriptsize $\pm$0.71}} & \makecell{45.58 \\ {\scriptsize $\pm$1.81}} & \makecell{45.03 \\ {\scriptsize $\pm$0.77}} & \makecell{45.93 \\ {\scriptsize $\pm$0.65}} & \makecell{46.08 \\ {\scriptsize $\pm$0.65}} & \makecell{45.68 \\ {\scriptsize $\pm$0.50}} \\
    \bottomrule
  \end{tabular}
  }
\end{table*}

\section{Optuna Hyper-Parameter Setup}

In our experiments, we employ the Optuna framework to automatically search for the optimal combination of hyper-parameters. 
As detailed in Table \cref{tab:ac_append,tab:med_append,tab:others_append,tab:rs_append}, we configured a total of 10 trials to explore the defined parameter space. Specifically, the learning rate (\texttt{lr}) and weight decay (\texttt{weight\_decay}) are sampled from a log-uniform distribution over the ranges  $[10^{-5}, 10^{-3}]$ and $[10^{-6}, 10^{-2}]$ respectively. The optimizer type (\texttt{optimtype}) is chosen from three candidates: AdamW, RMSprop, and Adam. For the encoder freezing parameter, \texttt{freeze\_encoders}, we generally set it to False by default. It is only included as a searchable categorical variable, allowing Optuna to decide between True and False, in specific experiments where we need to investigate its impact on model performance.
\section{Dataset Details}
For every dataset in MULTIBENCH++, we give a short overview that includes following items: (1) where the data come from and what they contain, (2) how we cleaned and extracted features following recent work, and(3) the exact train, validation, and test splits we use.
\begin{table*}[htbp]
  \centering
  \caption{Performance on Other Datasets. 
  Results are shown as \textit{mean} with the $\pm$\,\textit{standard deviation} value below it in a smaller font.
  The dash ``-" indicates the method is not applicable. }
  \label{tab:others_append}

  \renewcommand{\cellalign}{c} 
  \resizebox{\textwidth}{!}{
  \begin{tabular}{@{}l *{11}{c} @{}}
    \toprule
    \textbf{Dataset} & 
    \textbf{Concat} & \textbf{TF} & \textbf{ConcatEarly} & \textbf{LFT} & \textbf{EFT} & 
    \textbf{Multi-to-One} & \textbf{One-to-Multi} & \textbf{CAF} & \textbf{CACF} & \textbf{LS} & \textbf{TMC} \\
    \midrule
    MIRFLICKR & \makecell{62.81 \\ {\scriptsize $\pm$1.52}} & \makecell{62.33 \\ {\scriptsize $\pm$2.23}} & \makecell{62.42 \\ {\scriptsize $\pm$1.18}} & \makecell{50.25 \\ {\scriptsize $\pm$1.67}} & \makecell{38.02 \\ {\scriptsize $\pm$2.69}} & \makecell{58.44 \\ {\scriptsize $\pm$0.93}} & \makecell{59.95 \\ {\scriptsize $\pm$1.40}} & \makecell{59.13 \\ {\scriptsize $\pm$0.27}} & \makecell{62.51 \\ {\scriptsize $\pm$1.21}} & \makecell{62.45 \\ {\scriptsize $\pm$1.39}} & \makecell{62.00 \\ {\scriptsize $\pm$1.47}} \\
    NYUDv2 & \makecell{59.02 \\ {\scriptsize $\pm$4.24}} & \makecell{61.47 \\ {\scriptsize $\pm$4.05}} & \makecell{58.82 \\ {\scriptsize $\pm$4.58}} & \makecell{47.96 \\ {\scriptsize $\pm$8.51}} & \makecell{31.70 \\ {\scriptsize $\pm$6.24}} & \makecell{59.58 \\ {\scriptsize $\pm$3.83}} & \makecell{63.20 \\ {\scriptsize $\pm$5.58}} & \makecell{60.55 \\ {\scriptsize $\pm$3.09}} & \makecell{64.02 \\ {\scriptsize $\pm$5.26}} & \makecell{66.87 \\ {\scriptsize $\pm$3.52}} & \makecell{66.16 \\ {\scriptsize $\pm$1.95}} \\
    SUN-RGBD &  \makecell{60.28 \\ {\scriptsize $\pm$1.65}} & \makecell{59.43 \\ {\scriptsize $\pm$1.43}} & \makecell{60.83 \\ {\scriptsize $\pm$1.90}} & \makecell{45.22 \\ {\scriptsize $\pm$0.30}} & \makecell{31.61 \\ {\scriptsize $\pm$2.61}} & \makecell{53.52 \\ {\scriptsize $\pm$2.16}} & \makecell{56.97 \\ {\scriptsize $\pm$1.11}} & \makecell{53.53 \\ {\scriptsize $\pm$0.37}} & \makecell{58.14 \\ {\scriptsize $\pm$1.94}} & \makecell{60.78 \\ {\scriptsize $\pm$2.26}} & \makecell{59.00 \\ {\scriptsize $\pm$1.99}} \\
    MVSA-Single & \makecell{79.83 \\ {\scriptsize $\pm$0.87}} & \makecell{78.36 \\ {\scriptsize $\pm$2.91}} & \makecell{78.29 \\ {\scriptsize $\pm$1.31}} & \makecell{68.40 \\ {\scriptsize $\pm$7.86}} & \makecell{63.39 \\ {\scriptsize $\pm$9.19}} & \makecell{79.32 \\ {\scriptsize $\pm$0.60}} & \makecell{77.78 \\ {\scriptsize $\pm$1.05}} & \makecell{79.51 \\ {\scriptsize $\pm$1.50}} & \makecell{78.03 \\ {\scriptsize $\pm$3.56}} & \makecell{79.25 \\ {\scriptsize $\pm$2.14}} & \makecell{67.50 \\ {\scriptsize $\pm$10.49}} \\
    N-MNIST+N-TIDIGITS & \makecell{94.99 \\ {\scriptsize $\pm$0.20}} & \makecell{94.28 \\ {\scriptsize $\pm$0.13}} & \makecell{95.26 \\ {\scriptsize $\pm$0.13}} & \makecell{80.99 \\ {\scriptsize $\pm$1.07}} & \makecell{30.45 \\ {\scriptsize $\pm$4.91}} & \makecell{93.52 \\ {\scriptsize $\pm$0.41}} & \makecell{94.34 \\ {\scriptsize $\pm$0.77}} & \makecell{94.06 \\ {\scriptsize $\pm$0.54}} & \makecell{95.26 \\ {\scriptsize $\pm$0.27}} & \makecell{94.88 \\ {\scriptsize $\pm$0.41}} & \makecell{94.23 \\ {\scriptsize $\pm$0.34}} \\
    E-MNIST+EEG & \makecell{58.72 \\ {\scriptsize $\pm$5.12}} & \makecell{58.21 \\ {\scriptsize $\pm$1.92}} & \makecell{61.28 \\ {\scriptsize $\pm$3.22}} & \makecell{17.69 \\ {\scriptsize $\pm$3.22}} & \makecell{7.95 \\ {\scriptsize $\pm$0.36}} & \makecell{42.56 \\ {\scriptsize $\pm$5.66}} & \makecell{30.51 \\ {\scriptsize $\pm$9.02}} & \makecell{49.74 \\ {\scriptsize $\pm$3.10}} & \makecell{59.49 \\ {\scriptsize $\pm$3.10}} & \makecell{62.05 \\ {\scriptsize $\pm$1.58}} & \makecell{57.69 \\ {\scriptsize $\pm$3.14}} \\
    \bottomrule
  \end{tabular}
  }
\end{table*}

\begin{table*}[!htbp]
\centering
\caption{Optuna hyper-parameter search space (trials = 10)}
\begin{tabular}{lllc}
\toprule
\textbf{Parameter} & \textbf{Type} & \textbf{Sampler} & \textbf{Search Space / Candidates} \\
\midrule
\texttt{lr}              & Continuous (log) & \texttt{suggest\_loguniform} & $[10^{-5},\ 10^{-3}]$ \\
\texttt{weight\_decay}   & Continuous (log) & \texttt{suggest\_loguniform} & $[10^{-6},\ 10^{-2}]$ \\
\texttt{freeze\_encoders}& Categorical      & \texttt{suggest\_categorical} & \{\texttt{True},\texttt{False}\} \\
\texttt{optimtype}       & Categorical      & \texttt{suggest\_categorical} & \{\texttt{AdamW},\ \texttt{RMSprop},\ \texttt{Adam}\} \\
\bottomrule
\end{tabular}
\label{tab:optuna_space}
\end{table*}
\subsection*{MELD}
\paragraph{Data Source and Content} The Multimodal EmotionLines Dataset (MELD) is a large-scale dataset for emotion recognition in conversations. It contains over 13,000 utterances from the TV show \textit{Friends}, with audio, video, and text.
\paragraph{Link} \url{https://github.com/declare-lab/MELD}
\paragraph{Feature Extraction} 
\paragraph{Data Splits} The official split is used, containing 9989 training, 1109 validation, and 2610 test utterances.

\subsection*{IEMOCAP}
\paragraph{Data Source and Content} The Interactive Emotional Dyadic Motion Capture (IEMOCAP) database is a popular dataset for emotion recognition, containing approximately 12 hours of audiovisual data from ten actors.
\paragraph{Link} \url{https://sail.usc.edu/iemocap/}
\paragraph{Feature Extraction} 
We follow \url{https://github.com/soujanyaporia/multimodal-sentiment-analysis/tree/master?tab=readme-ov-file} and preprocess the features.
\paragraph{Data Splits} The official split is used, containing 5810 training and 1623 test utterances. Since the official dataset split does not include a dedicated validation set, we reuse the test set as the validation set.

\subsection*{MAMI}
\paragraph{Data Source and Content} A dataset for Misogynistic Meme Detection, containing over 10,000 image-text memes annotated for misogyny and other categories.
\paragraph{Link} \url{https://github.com/MIND-Lab/SemEval2022-Task-5-Multimedia-Automatic-Misogyny-Identification-MAMI-}
\paragraph{Feature Extraction} A \texttt{ResNet} is used to encode the image component, and a \texttt{BERT} model is used to encode the text.
\paragraph{Data Splits} The official split is used, containing 9000 training, 1000 validation, and 1000 test utterances.

\subsection*{Memotion}
\paragraph{Data Source and Content} A dataset for analyzing emotions in memes. It contains 10,000 memes annotated for sentiment and three types of emotions (humor, sarcasm, motivation).
\paragraph{Link} \url{https://www.kaggle.com/datasets/williamscott701/memotion-dataset-7k}
\paragraph{Feature Extraction} A \texttt{ResNet} is used to encode the image component, and a \texttt{BERT} model is used to encode the text.
\paragraph{Data Splits} We adopt an 8:1:1 split for training, validation, and test sets, containing 5465 training, 683 validation, and 683 test utterances, with the random seed fixed at 42.

\subsection*{MUTE}
\paragraph{Data Source and Content} This datasets focus on detecting harmful content. MUTE targets troll-like behavior in image/text posts, while MultiOFF focuses on identifying offensive content and its target.
\paragraph{Link} \url{https://github.com/eftekhar-hossain/MUTE-AACL22}
\paragraph{Feature Extraction} A \texttt{ResNet} is used to encode the image component, and a \texttt{BERT} model is used to encode the text.
\paragraph{Data Splits} The official split is used, containing 3365 training, 375 validation, and 416 test utterances.

\subsection*{MultiOFF}
\paragraph{Data Source and Content} This datasets focus on detecting harmful content. MUTE targets troll-like behavior in image/text posts, while MultiOFF focuses on identifying offensive content and its target.
\paragraph{Link} \url{https://github.com/bharathichezhiyan/Multimodal-Meme-Classification-Identifying-Offensive-Content-in-Image-and-Text}
\paragraph{Feature Extraction} A \texttt{ResNet} is used to encode the image component, and a \texttt{BERT} model is used to encode the text.
\paragraph{Data Splits} The official split is used, containing 445 training, 149 validation, and 149 test utterances.

\subsection*{MET-Meme (Chinese)}
\paragraph{Data Source and Content} A dataset for multimodal metaphor detection in memes, available in both English and Chinese versions.
\paragraph{Link} \url{https://github.com/liaolianfoka/MET-Meme-A-Multi-modal-Meme-Dataset-Rich-in-Metaphors}
\paragraph{Feature Extraction} A \texttt{ResNet} is used to encode the image component, and a \texttt{BERT} model is used to encode the text.
\paragraph{Data Splits} We adopt an 7:1:2 split for training, validation, and test sets, containing 1,609 training, 229 validation, and 461 test utterances, with the random seed fixed at 42.

\subsection*{MET-Meme (English)}
\paragraph{Data Source and Content} A dataset for multimodal metaphor detection in memes, available in both English and Chinese versions.
\paragraph{Link} \url{https://github.com/liaolianfoka/MET-Meme-A-Multi-modal-Meme-Dataset-Rich-in-Metaphors}
\paragraph{Feature Extraction} A \texttt{ResNet} is used to encode the image component, and a \texttt{BERT} model is used to encode the text.
\paragraph{Data Splits} We adopt an 7:1:2 split for training, validation, and test sets, containing 737 training, 105 validation, and 211 test utterances, with the random seed fixed at 42.

\subsection*{CH-SIMS}
\paragraph{Data Source and Content} A fine-grained single- and multi-modal sentiment analysis dataset in Chinese. It contains over 2,200 short videos with unimodal and multimodal annotations.
\paragraph{Link} \url{https://github.com/thuiar/MMSA}
\paragraph{Feature Extraction} We use the pre-extracted features provided by the dataset without any additional modifications.
\paragraph{Data Splits} The official split is used, containing 1368 training, 456 validation, and 457 test utterances.

\subsection*{CH-SIMSv2}
\paragraph{Data Source and Content} A fine-grained single- and multi-modal sentiment analysis dataset in Chinese. It contains over 2,200 short videos with unimodal and multimodal annotations.
\paragraph{Link} \url{https://thuiar.github.io/sims.github.io/chsims}
\paragraph{Feature Extraction} We use the pre-extracted features provided by the dataset without any additional modifications.
\paragraph{Data Splits} The official split is used, containing 2722 training, 647 validation, and 1034 test utterances.

\subsection*{Twitter2015}
\paragraph{Data Source and Content} Originating from SemEval tasks, these datasets were extended for multimodal aspect-based sentiment analysis, pairing tweets with relevant images.
\paragraph{Link} \url{https://archive.org/details/twitterstream}
\paragraph{Feature Extraction} A \texttt{ResNet} is used to encode the image component, and a \texttt{BERT} model is used to encode the text.
\paragraph{Data Splits} The official split is used, containing 3179 training, 1122 validation, and 1037 test utterances.

\subsection*{Twitter1517}
\paragraph{Data Source and Content} 
\paragraph{Link} \url{https://github.com/code-chendl/HFIR}
\paragraph{Feature Extraction} A \texttt{ResNet} is used to encode the image component, and a \texttt{BERT} model is used to encode the text.
\paragraph{Data Splits} We adopt an 7:1:2 split for training, validation, and test sets, containing 3270 training, 467 validation, and 935 test utterances, with the random seed fixed at 42.

\subsection*{Houston2013}
\paragraph{Data Source and Content} The 2013 dataset, from the IEEE GRSS Data Fusion Contest, provides HSI and LiDAR data over the University of Houston campus, covering 15 land use classes. The 2018 version is a more complex dataset covering 20 urban land use classes.
\paragraph{Link} \url{https://machinelearning.ee.uh.edu/?page_id=459}
\paragraph{Feature Extraction} Following the link \url{https://github.com/songyz2019/rs-fusion-datasets}, we will proceed with the direct acquisition and loading of the dataset, which will be used for the joint classification of hyperspectral, LiDAR, and SAR data. The HSI data is processed by the \texttt{conv\_hsi} encoder, and the LiDAR data is processed by the \texttt{conv\_dsm} encoder.
\paragraph{Data Splits} The official split is used, containing 2817 training and 12182 test utterances. Since the official dataset split does not include a dedicated validation set, we reuse the test set as the validation set.

\subsection*{Houston2018}
\paragraph{Data Source and Content} The 2013 dataset, from the IEEE GRSS Data Fusion Contest, provides HSI and LiDAR data over the University of Houston campus, covering 15 land use classes. The 2018 version is a more complex dataset covering 20 urban land use classes.
\paragraph{Link} \url{https://machinelearning.ee.uh.edu/2018-ieee-grss-data-fusion-challenge-fusion-of-multispectral-lidar-and-hyperspectral-data/}
\paragraph{Feature Extraction} Following the link \url{https://github.com/songyz2019/rs-fusion-datasets}, we will proceed with the direct acquisition and loading of the dataset, which will be used for the joint classification of hyperspectral, LiDAR, and SAR data. The HSI data is processed by the \texttt{conv\_hsi} encoder, and the LiDAR data is processed by the \texttt{conv\_dsm} encoder.
\paragraph{Data Splits} The official split is used, containing 18750 training and 2000160 test utterances. Since the official dataset split does not include a dedicated validation set, we reuse the test set as the validation set.

\subsection*{MUUFL Gulfport}
\paragraph{Data Source and Content} The MUUFL dataset contains HSI and LiDAR data collected over the University of Southern Mississippi, Gulfport campus. The ground truth contains 11 urban land use classes.
\paragraph{Link} \url{https://github.com/GatorSense/MUUFLGulfport}
\paragraph{Feature Extraction} Following the link \url{https://github.com/songyz2019/rs-fusion-datasets}, we will proceed with the direct acquisition and loading of the dataset, which will be used for the joint classification of hyperspectral, LiDAR, and SAR data. We use the \texttt{conv\_hsi} encoder for the HSI data and the \texttt{conv\_dsm} encoder for the LiDAR data.
\paragraph{Data Splits} The official split is used, containing 1100 training and 52587 test utterances. Since the official dataset split does not include a dedicated validation set, we reuse the test set as the validation set.

\subsection*{Trento}
\paragraph{Data Source and Content} This dataset covers a rural area south of Trento, Italy. It combines HSI data with corresponding LiDAR-derived Digital Surface Model (DSM) data, with 6 classified land cover classes.
\paragraph{Link} \url{https://github.com/tyust-dayu/Trento/tree/b4afc449ce5d6936ddc04fe267d86f9f35536afd}
\paragraph{Feature Extraction} Following the link \url{https://github.com/songyz2019/rs-fusion-datasets}, we will proceed with the direct acquisition and loading of the dataset, which will be used for the joint classification of hyperspectral, LiDAR, and SAR data. The HSI data is processed by \texttt{conv\_hsi}, and the LiDAR data by \texttt{conv\_dsm}.
\paragraph{Data Splits} The official split is used, containing 600 training and 29614 test utterances. Since the official dataset split does not include a dedicated validation set, we reuse the test set as the validation set.

\subsection*{Berlin}
\paragraph{Data Source and Content} This dataset provides co-registered HSI and Synthetic Aperture Radar (SAR) data for the city of Berlin, Germany, with 8 distinct urban land cover classes.
\paragraph{Link} \url{https://gfzpublic.gfz-potsdam.de/pubman/faces/ViewItemFullPage.jsp?itemId=item_1480927_5}
\paragraph{Feature Extraction} Following the link \url{https://github.com/songyz2019/rs-fusion-datasets}, we will proceed with the direct acquisition and loading of the dataset, which will be used for the joint classification of hyperspectral, LiDAR, and SAR data. The HSI data is processed by \texttt{conv\_hsi}. The SAR data, being image-like, is processed by a \texttt{ResNet} encoder.
\paragraph{Data Splits} The official split is used, containing 2820 training and 461851 test utterances. Since the official dataset split does not include a dedicated validation set, we reuse the test set as the validation set.

\subsection*{MDAS (Augsburg)}
\paragraph{Data Source and Content} This dataset contains multi-sensor data for urban area classification in Augsburg, Germany, featuring HSI and SAR imagery.
\paragraph{Link} \url{https://github.com/songyz2019/rs-fusion-datasets}
\paragraph{Feature Extraction} Following the link \url{https://github.com/songyz2019/rs-fusion-datasets}, we will proceed with the direct acquisition and loading of the dataset, which will be used for the joint classification of hyperspectral, LiDAR, and SAR data. The HSI data uses the \texttt{conv\_hsi} encoder, while the SAR data is processed using a \texttt{ResNet} encoder.
\paragraph{Data Splits} The official split is used, containing 761 training and 77533 test utterances. Since the official dataset split does not include a dedicated validation set, we reuse the test set as the validation set.

\subsection*{ForestNet}
\paragraph{Data Source and Content} A dataset for wildfire prevention, containing satellite imagery from Sentinel-2, topography data (DSM), and weather data. It's used to predict wildfire risk.
\paragraph{Link} \url{https://github.com/spott/ForestNet}
\paragraph{Feature Extraction} A \texttt{ResNet} processes satellite imagery, \texttt{ResNet} processes topography data, and an MLP processes tabular weather data (if needed).
\paragraph{Data Splits} The official split is used, containing 1616 training, 473 validation, and 668 test utterances.

\subsection*{TCGA-BRCA}
\paragraph{Data Source and Content} Data from The Cancer Genome Atlas (TCGA) Program, focusing on Breast Cancer (BRCA). It's a multi-omics dataset, often including gene expression, DNA methylation, and copy number variation, used for tasks like survival prediction.
\paragraph{Link} \url{https://github.com/txWang/MOGONET}
\paragraph{Feature Extraction} Each tabular omics modality is encoded via an independent fully-connected linear layer.
\paragraph{Data Splits} The dataset is extracted sequentially: the leading 20\% forms the test set, the subsequent 5\% the validation set, and the remaining 75\% the training set, containing 657 training, 43 validation, and 175 test utterances.

\subsection*{TCGA}
\paragraph{Data Source and Content} Data from The Cancer Genome Atlas (TCGA) Program, focusing on Breast Cancer (BRCA). It's a multi-omics dataset, often including gene expression, DNA methylation, and copy number variation, used for tasks like survival prediction.
\paragraph{Link} \url{https://www.cancer.gov/ccg/access-data}
\paragraph{Feature Extraction}We use the TCGA dataset selected by \url{https://github.com/bowang-lab/IntegrAO}. Each tabular omics modality is encoded via an independent fully-connected linear layer.
\paragraph{Data Splits} The dataset is extracted sequentially: the first 60\% for training, the next 25\% for validation, and the final 20\% for testing, containing 169 training, 76 validation, and 61 test utterances.

\subsection*{ROSMAP}
\paragraph{Data Source and Content} Data from the Religious Orders Study and Memory and Aging Project (ROSMAP) for Alzheimer's disease research. It includes multi-omics data from post-mortem brain tissue.
\paragraph{Link} \url{https://github.com/txWang/MOGONET}
\paragraph{Feature Extraction} Each tabular omics modality is encoded via an independent Identity mapping encoder.
\paragraph{Data Splits} The dataset is extracted sequentially: the first 60\% for training, the next 25\% for validation, and the final 20\% for testing, containing 194 training, 87 validation, and 70 test utterances.

\subsection*{SIIM-ISIC}
\paragraph{Data Source and Content} From the 2020 SIIM-ISIC Melanoma Classification Kaggle challenge. The dataset contains thousands of dermoscopic images of skin lesions, with patient-level metadata, for classifying lesions.
\paragraph{Link} \url{https://www.kaggle.com/competitions/siim-isic-melanoma-classification/data}
\paragraph{Feature Extraction} A \texttt{ResNet} encoder is used for the dermoscopic images, and an independent Identity mapping encoder is used for the tabular patient metadata.
\paragraph{Data Splits} We adopt an 8:1:1 split for training, validation, and test sets, containing 26502 training, 3312 validation, and 3312 test utterances, with the random seed fixed at 42.

\subsection*{Derm7pt}
\paragraph{Data Source and Content} A multiclass skin lesion classification dataset based on the 7-point checklist. It provides dermoscopic images and a corresponding vector of semi-quantitative clinical features.
\paragraph{Link} \url{https://www.kaggle.com/datasets/menakamohanakumar/derm7pt}
\paragraph{Feature Extraction} The images are encoded with a CNNEncoder, while the tabular clinical feature vectors are encoded with Identity mapping encoder.
\paragraph{Data Splits} The official split is used, containing 413 training, 203 validation, and 395 test utterances.

\subsection*{GAMMA}
\paragraph{Data Source and Content} From the Glaucoma grAding from Multi-Modality imAges challenge. It contains color fundus images and stereo-pairs of disc photos for glaucoma diagnosis.
\paragraph{Link} \url{https://zenodo.org/records/15119049}
\paragraph{Feature Extraction} Both the fundus images and stereo-pair images are encoded using a CNNEncoder.
\paragraph{Data Splits} The dataset is extracted sequentially: the first 20\% for training, the next 10\% for validation, and the final 70\% for testing, containing 20 training, 10 validation, and 70 test utterances.

\subsection*{MIMIC-III}
\paragraph{Data Source and Content} A large, de-identified ICU database containing structured data (lab results, vitals) and unstructured clinical notes for over 40,000 patients. It's used for tasks like mortality prediction.
\paragraph{Link} \url{https://physionet.org/content/mimiciii/1.4/}
\paragraph{Feature Extraction} Modality is encoded by a TimeSeriesTransformerEncoder  and Identity mapping.
\paragraph{Data Splits} The dataset is extracted sequentially: the leading 20\% forms the test set, the subsequent 5\% the validation set, and the remaining 75\% the training set, containing 24462 training, 1631 validation, and 6523 test utterances.

\subsection*{MIMIC-CXR}
\paragraph{Data Source and Content} A large-scale dataset containing over 377,000 chest X-ray images and their corresponding free-text radiology reports.
\paragraph{Link} \url{https://physionet.org/content/mimic-cxr-jpg/2.1.0/}
\paragraph{Feature Extraction} We use a \texttt{ResNet} encoder for the chest X-ray images and a \texttt{BERT} model for the radiology reports.
\paragraph{Data Splits} Following the official split, we cap the training set to a maximum of 5,000 utterances, yielding 5,000 for training, 2942 for validation, and 5117 for test. We set the random seed=42 to select the training samples. 

\subsection*{eICU}
\paragraph{Data Source and Content} A multi-center ICU database with de-identified data for over 200,000 admissions from hospitals across the US. It contains high-granularity vital sign data and clinical notes.
\paragraph{Link} \url{https://eicu-crd.mit.edu/https://physionet.org/content/eicu-crd/2.0/}
\paragraph{Feature Extraction} All modalities are encoded using the identity mapping.
\paragraph{Data Splits} The corpus is hierarchically partitioned: an initial 80 / 20 split isolates the test set; the remaining 80\% is then subdivided at a 15 : 1 ratio, producing final allocations of approximately 75\% training, 5\% validation, and 20\% test. We then get 5727 training, 382 validation, and 1528 test utterances.The random seed is fixed at 42.

\subsection*{MIRFLICKR}
\paragraph{Data Source and Content} A dataset of one million images from the Flickr website with their associated user-assigned tags. A 25,000-image subset with curated labels is commonly used for multimodal retrieval and classification.
\paragraph{Link} \url{https://press.liacs.nl/mirflickr/}
\paragraph{Feature Extraction} A \texttt{ResNet} is used for the images, and a \texttt{BERT} model is used for the textual tags.
\paragraph{Data Splits} We adopt an 7:1:2 split for training, validation, and test sets, containing 14010 training, 2001 validation, and 4004 test utterances, with the random seed fixed at 42.

\subsection*{CUB Image-Caption}
\paragraph{Data Source and Content} An extension of the Caltech-UCSD Birds-200-2011 dataset. It pairs detailed bird images with rich, descriptive text captions, ideal for fine-grained image-text matching.
\paragraph{Link} \url{https://github.com/iffsid/mmvae}
\paragraph{Feature Extraction} The bird images are encoded with a \texttt{ResNet}, and the text captions are encoded with a \texttt{BERT} model.
\paragraph{Data Splits} We adopt an 7:1.5:1.5 split for training, validation, and test sets, containing 82510 training, 17680 validation, and 17690 test utterances, with the random seed fixed at 2025.

\subsection*{SUN-RGBD}
\paragraph{Data Source and Content} A large-scale dataset for indoor scene understanding. It contains over 10,000 RGB-D (color and depth) images with dense annotations.
\paragraph{Link} \url{https://rgbd.cs.princeton.edu/}
\paragraph{Feature Extraction}We use a \texttt{ResNet} encoder for the RGB images and a similar \texttt{ResNet} architecture for the depth maps.
\paragraph{Data Splits} The official split is used, containing 4845 training and 4659 test utterances. Since the official dataset split does not include a dedicated validation set, we reuse the test set as the validation set.

\subsection*{NYUDv2}
\paragraph{Data Source and Content} The NYU-Depth Dataset V2 provides RGB and Depth images of various indoor scenes captured from a Microsoft Kinect, with dense labels for semantic segmentation.
\paragraph{Link} \url{https://cs.nyu.edu/~fergus/datasets/nyu_depth_v2.html}
\paragraph{Feature Extraction} A \texttt{ResNet} encoder is used for the RGB images and another for the depth images.
\paragraph{Data Splits} The official split is used, containing 795 training, 414 validation, and 654 test utterances. 

\subsection*{UPMC-Food101}
\paragraph{Data Source and Content} An extension of Food-101, this dataset pairs food images with their ingredient lists for tasks like recipe retrieval from images.
\paragraph{Link} \url{https://www.kaggle.com/datasets/gianmarco96/upmcfood101}
\paragraph{Feature Extraction}  We follow \url{https://github.com/facebookresearch/mmbt} to prepare the splits. A \texttt{ResNet} encodes the food images, while a \texttt{BERT} model encodes the ingredient lists.
\paragraph{Data Splits} The official split is used, containing 62971 training, 5000 validation, and 22715 test utterances. 

\subsection*{MVSA-Single}
\paragraph{Data Source and Content} A Multi-View Sentiment Analysis dataset containing image-text posts from Twitter. The ``Single" variant contains posts where the sentiment label is consistent across annotators.
\paragraph{Link} \url{https://www.kaggle.com/datasets/vincemarcs/mvsasingle}
\paragraph{Feature Extraction}  We follow \url{https://github.com/facebookresearch/mmbt} to prepare the splits. Images are encoded using \texttt{ResNet}, and the tweet text is encoded using \texttt{BERT}.
\paragraph{Data Splits} The official split is used, containing 1555 training, 518 validation, and 519 test utterances. 

\subsection*{MNIST-SVHN}
\paragraph{Data Source and Content} This is a synthetic dataset combining two famous digit recognition datasets: MNIST (handwritten digits) and SVHN (street view house numbers). The task is to classify pairs of digits.
\paragraph{Link} \url{https://github.com/iffsid/mmvae}
\paragraph{Feature Extraction} We apply a flattening operation to MNIST images, whereas SVHN images are processed by a simple CNN encoder.
\paragraph{Data Splits} The official split is used, containing 560,680 training and 100,000 test utterances. Since the official dataset split does not include a dedicated validation set, we reuse the test set as the validation set.

\subsection*{N-MNIST+N-TIDIGITS}
\paragraph{Data Source and Content} Neuromorphic versions of MNIST (vision) and TIDIGITS (audio-spoken digits). Data is recorded as asynchronous event streams (spikes) from event-based sensors.
\paragraph{Feature Extraction} We follow \url{https://github.com/MrLinNing/MemristorLSM}, and pair each N-MNIST frame with its corresponding N-TIDIGITS audio clip to form unique image–sound pairs and perform classification on these aligned inputs. NMNIST frames are encoded by a CNN; NTIDIGITS MFCCs are encoded by an LSTM.
\paragraph{Data Splits} The dataset is extracted sequentially: the first 70\% for training, the next 15\% for validation, and the final 15\% for testing, containing 2835 training, 603 validation, and 612 test utterances.

\subsection*{E-MNIST+EEG}
\paragraph{Data Source and Content} A dataset that combines images from the E-MNIST dataset (handwritten letters/digits) with simultaneously recorded EEG brain signals from subjects viewing them.
\paragraph{Feature Extraction} We follow \url{https://github.com/MrLinNing/MemristorLSM}, and construct unique image–EEG pairs by pairing each E-MNIST sample with its corresponding EEG recording, and then perform classification on these paired inputs. The E-MNIST images are encoded with a CNN, and the time-series EEG signals are encoded with an LSTM encoder.
\paragraph{Data Splits} The dataset is extracted sequentially: the first 70\% for training, the next 15\% for validation, and the final 15\% for testing, containing 468 training, 104 validation, and 130 test utterances.


\end{document}